\documentclass[journal]{IEEEtran}

\ifCLASSINFOpdf
\else
\fi

\RequirePackage{amsmath}
\usepackage{amsmath,amsfonts,amssymb}
\usepackage{algpseudocode}
\usepackage{algorithm}
\usepackage{algorithmicx}
\usepackage{graphicx}
\usepackage{siunitx}
\usepackage{cite}
\usepackage{textcomp}
\usepackage{gensymb}
\DeclareMathOperator*{\argmin}{argmin}
\usepackage{boldline,multirow}
\usepackage{color}
\usepackage{soul} 
\usepackage[table]{xcolor}
\newcommand{\Reals}{\rm I\!R}
\newcommand{\tabincell}[2]{\begin{tabular}{@{}#1@{}}#2\end{tabular}}
\usepackage{nameref,zref-xr}
\usepackage[colorlinks=true,citecolor=blue,urlcolor=black]{hyperref}

\begin{document}

\title{Context Label Learning: Improving Background Class Representations in Semantic Segmentation}

\author{Zeju~Li,
        Konstantinos~Kamnitsas,
        Cheng~Ouyang,
        Chen~Chen
        and~Ben~Glocker
\thanks{Z. Li, K. Kamnitsas, C. Ouyang, C. Chen and B. Glocker are with the BioMedIA Group, Department of Computing, Imperial College London, SW7 2AZ, United Kingdom. K. Kamnitsas is also with School of Computer Science, University of Birmingham, B15 2SQ, United Kingdom and Department of Engineering Science, University of Oxford, OX1 3PJ, United Kingdom. C.Chen was also with HeartFlow, Inc., USA. E-mail: zeju.li18@imperial.ac.uk.}
}


\maketitle

\begin{abstract}

Background samples provide key contextual information for segmenting regions of interest (ROIs). However, they always cover a diverse set of structures, causing difficulties for the segmentation model to learn good decision boundaries with high sensitivity and precision. The issue concerns the highly heterogeneous nature of the background class, resulting in multi-modal distributions. Empirically, we find that neural networks trained with heterogeneous background struggle to map the corresponding contextual samples to compact clusters in feature space. As a result, the distribution over background logit activations may shift across the decision boundary, leading to systematic over-segmentation across different datasets and tasks. In this study, we propose context label learning (CoLab) to improve the context representations by decomposing the background class into several subclasses. Specifically, we train an auxiliary network as a task generator, along with the primary segmentation model, to automatically generate context labels that positively affect the ROI segmentation accuracy. Extensive experiments are conducted on several challenging segmentation tasks and datasets. The results demonstrate that CoLab can guide the segmentation model to map the logits of background samples away from the decision boundary, resulting in significantly improved segmentation accuracy. Code is available\footnote{\url{https://github.com/ZerojumpLine/CoLab}}.

\end{abstract}

\begin{IEEEkeywords}
underfitting, multi-task learning, self-supervised learning, image segmentation.
\end{IEEEkeywords}

%

\IEEEpeerreviewmaketitle

\section{Introduction}
\label{sec:introduction}

\begin{figure}[t]
\centering
\includegraphics[width=0.48\textwidth]{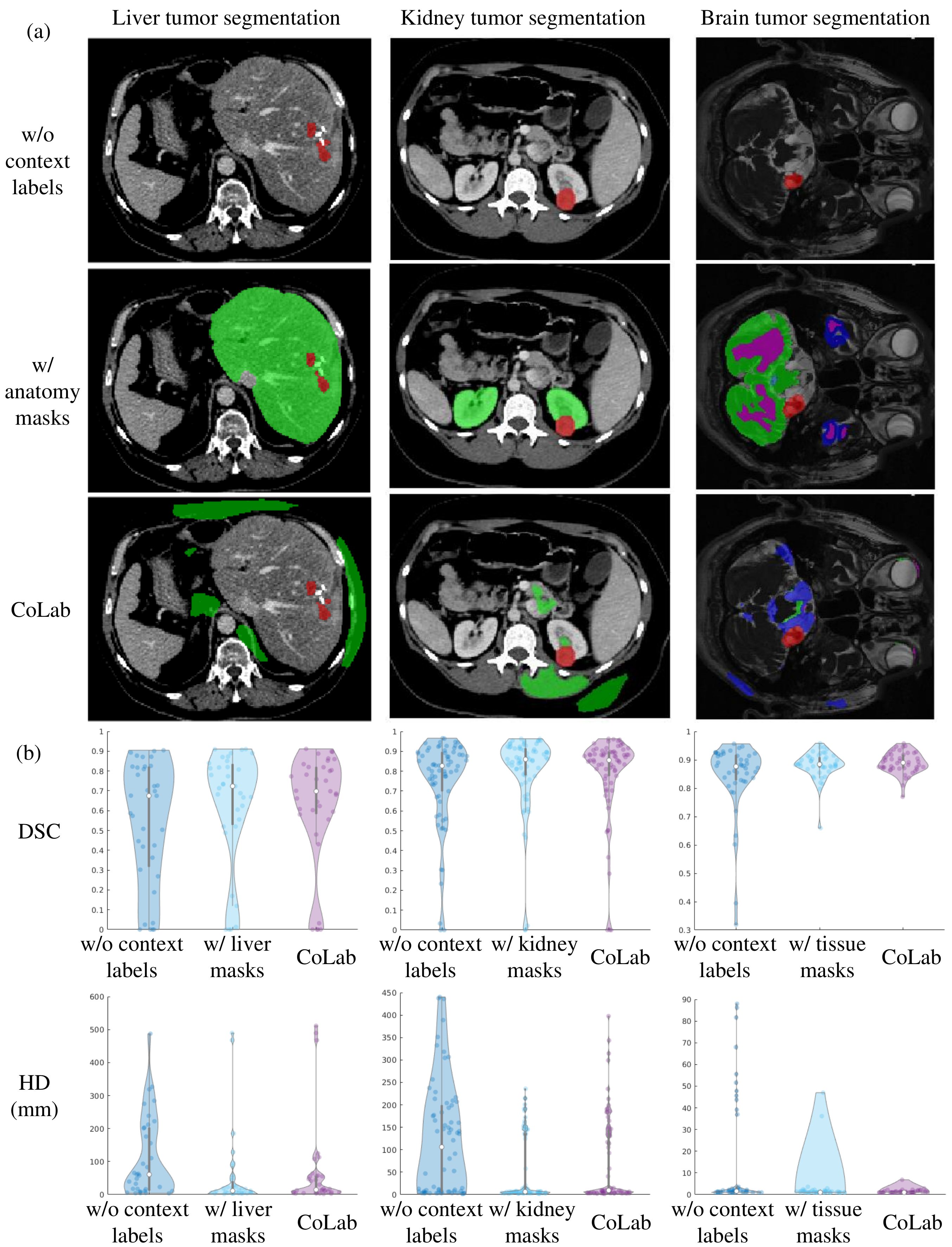}

\caption{(a) Examples of ROI classes (in red color) along with human-defined context labels and context labels generated automatically by the proposed context label learning (CoLab) in background. Here, human-defined context labels (e.g., liver, kidney, brain tissue) provide additional anatomical information to benefit representation learning. In a similar manner, CoLab automatically finds and labels task-related tissues/structures in background to benefit corresponding ROI segmentation, see Fig.~\ref{fig_contexttask} for more details. (b) Self-supervised context labels generated by CoLab can significantly improve the segmentation of different structures with improved Dice similarity coefficient (DSC) and reduced 95\% Hausdorff distance (HD), compared to the one trained w/o context labels. The performance gains are comparable and sometimes even higher than the ones using human-defined context labels.} \label{fig_mainidea}
\end{figure}

\IEEEPARstart{C}{onvolutional} neural networks (CNNs) are state-of-the-art approach for semantic image segmentation. Their large number of trainable parameters make them capable to fit to different kinds of tasks yielding high performance in terms segmentation accuracy~\cite{zhang2021understanding}. Yet, in real world applications, CNNs seem sometimes unable to capture and generalize from complex, heterogeneous training sets when the amount of training data is limited~\cite{novak2018sensitivity}. Specifically, in the case of medical image segmentation, samples from the background class, which provide essential contextual information for regions of interest (ROIs), make up the majority of the training set while containing diverse sets of structures with heterogeneous characteristics, making it hard for the segmentation model to learned accurate decision boundaries.

When trained with datasets with a highly heterogeneous background class, the segmentation model is prone to underfit these contextual samples and fail to separate the ones which share characteristics similar to the ROI samples. The model then produces false positives (FP), yielding systematic over-segmentation. In this study, we argue that underfitting of the contextual information is a main cause of degraded segmentation performance by affecting precision (calculated as $\frac{\text{TP}}{\text{TP}+\text{FP}}$). We observe that better context representation where the background class is decomposed into several subclasses, for example, using additional anatomy labels, significantly improves the ROI segmentation accuracy. We show some examples of human-defined context labels in Fig.~\ref{fig_mainidea}(a). In the case of liver tumor segmentation, for example, it is beneficial to also have labels for the liver available in addition to the tumor class. Empirically, we find that when training with human-defined context labels, the segmentation model can yield better performance in terms of Dice similarity coefficient (DSC) and 95\% Hausdorff distance (HD), as shown in Fig.~\ref{fig_mainidea}(b). However, these human-defined context labels are not always available and difficult and time-consuming to obtain. In many applications, only the ROI labels, e.g., tumor class, are available. Here, we propose context label learning (CoLab), which automatically generates context labels to improve the learning of a context representation yielding better ROI segmentation accuracy. As demonstrated in Fig.~\ref{fig_mainidea}(b), CoLab can bring similar improvements when compared with training with human-defined context labels, without the need for expert knowledge.

The contribution of this study can be summarized as follows: 1) With the observations of six datasets, we conclude that underfitting of the background class consistently degrades the segmentation performance by decreasing precision. 2) We find that better context representations with a decomposition of the background class can improve segmentation performance. 3) We propose CoLab, a flexible and generic method to automatically generate soft context labels. We validate CoLab with extensive experiments and find consistent improvements where the segmentation accuracy is en par and sometimes better compared to the case where human annotated context labels are available.

\section{Related work}
\label{sec:relatedwork}

\subsection{Class imbalance}

CoLab is related to the class imbalance problem as background commonly constitutes the majority class in image segmentation. However, methods to combat class imbalance mostly focus explicitly on improving the performance of the minority categories~\cite{kang2019decoupling, xiang2020learning, liu2019large}. Most approaches ignore the characteristics of majority background class as it is not contributing much to the common evaluation metrics of sensitivity, precision, and DSC of the ROI classes. CoLab focuses specifically on the representation of the background class and is complementary to methods tackling class imbalance such as loss reweighting strategies~\cite{cai2021ace}.

Previous studies adopt coarse-to-fine strategies to reduce FP in segmentation tasks with class imbalance~\cite{setio2016pulmonary, valindria2018small, zhou2017fixed}. However, ROI samples missed in the coarse stage cannot be recovered with later stages. In contrast, CoLab is trained in an end-to-end manner and can reduce varied kinds of FP.

\subsection{Multi-task learning}

CoLab, which is formulated as multi-label classification, can be seen as a form of multi-task learning (MTL). Current MTL methods train the model with different predefined tasks together with the main task using a shared feature representation~\cite{li2019deepvolume, qaiser2021multiple, bragman2019stochastic}. Previous works also attempted to incorporate spatial prior~\cite{zhang2021automatic, glocker2012joint} or task prior~\cite{zhang2020exploring, zhou2019prior} into model training with some predefined auxiliary tasks and optimization functions. In contrast, CoLab reformulates the main task by decomposing the background class with context labels and automatically generate the auxiliary task in a self-supervised manner. We argue that CoLab can have a direct impact on the main task by extending the label space.

The main methodology of the CoLab strategy is inspired by some recently proposed methods which aim to generate the weights for pre-defined auxiliary tasks or labels through a similar meta-learning framework~\cite{navon2020auxiliary, liu2019self}. In this study, CoLab is specifically designed for semantic segmentation with heterogeneous background classes which is a common scenario in medical imaging.

\section{Context labels in image segmentation}
\label{sec:analysis}

\begin{figure*}[t]
\centering
\includegraphics[width=\textwidth]{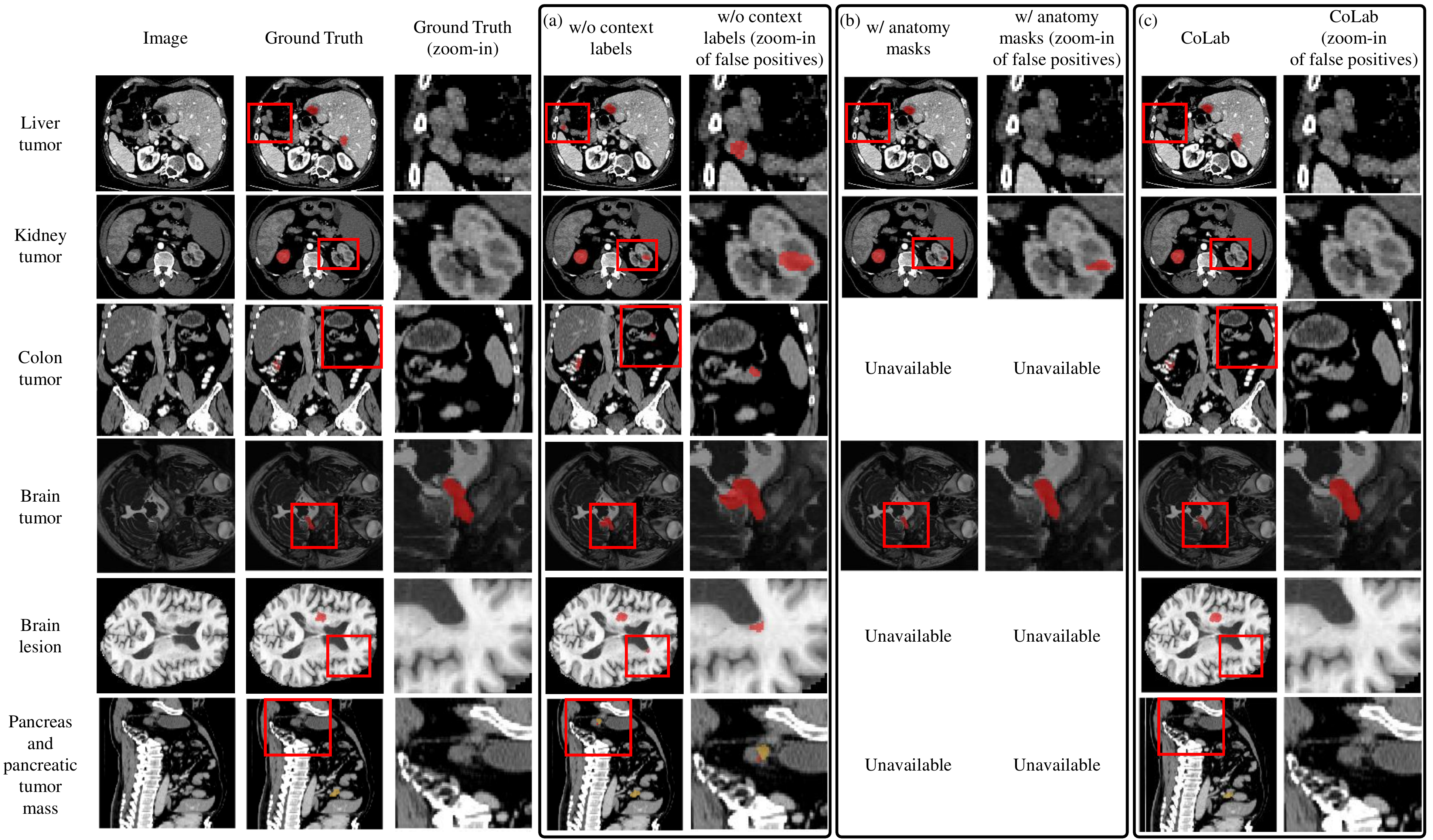}
 
\caption{Visualization of different datasets and segmentation results on test data when training w/o and w/ different types of context labels. (a) Models trained without any context labels are prone to over-segmentation of the ROI (red/red + orange) with different kinds of FP (marked with red boxes). (b) Empirically, we observe that adding human-defined context labels such as anatomy masks for supervision helps reduce FP and therefore obtains higher overall segmentation accuracy. (c) The proposed CoLab achieves a similar benefit without requiring additional human annotation by producing context labels automatically. } \label{fig_segexample}
\end{figure*}

\subsection{Preliminaries}

We consider CNNs for multi-class segmentation with a total of $c$ classes. Given a training dataset $\mathcal{D} = \{ (\boldsymbol{x}_i, \boldsymbol{y}_i) \}_{i = 1}^N$ with $N$ samples, where $\boldsymbol{y}_i \in \Reals^c$ is the one-hot encoded label for the central pixel in the image sample $\boldsymbol{x}_i \in \Reals^d$, such that $\boldsymbol{1} \cdot \boldsymbol{y}_i = 1$ $\forall$ $i$. A segmentation model $f_\phi(\cdot)$ learns class representations of the input sample $\boldsymbol{x}_i$, noted as $\boldsymbol{z}_i = f_\phi(\boldsymbol{x}_i) \in \Reals^c$. We obtain the predicted probability $\boldsymbol{p}_i$ that the real class of $\boldsymbol{x}_i$ is $j$ via a softmax function with $p_{ij} = \mathrm{e}^{z_{ij}} / \sum_{j=1}^c \mathrm{e}^{z_{ij}}$. Typically the model is optimized by minimizing the empirical risk $R_{\mathcal{L}_{seg}}(f_\phi)=\frac{1}{N}\sum_{i=1}^{N}\mathcal{L}_{seg}(f_\phi(\boldsymbol{x}_i),\boldsymbol{y}_i)$ computed on the training set. The segmentation loss $\mathcal{L}_{seg}$ can be defined as the sum of losses $\mathcal{L}$ over $c$ classes:

\begin{equation}
    \begin{split}
    & \mathcal{L}_{seg}(f_\phi(\boldsymbol{x}_i),\boldsymbol{y}_i) = \sum_{j=1}^{c} \mathcal{L}(p_{ij}, y_{ij}) \\
    & = \underbrace{\sum_{j=1}^{c-1}\mathcal{L}(p_{ij}, y_{ij})}_{\text{ROI classes}} + \underbrace{\mathcal{L}(p_{ic}, y_{ic})}_{\text{background class}},
    \label{eq:CE}
    \end{split}
\end{equation}

where $\mathcal{L}$ is a criterion for a specific class, such as cross entropy (CE) or soft DSC~\cite{milletari2016v}. Here, we further decompose $\mathcal{L}_{seg}$ into two terms, including an ROI loss (computed on $c-1$ foreground classes) and a background loss (computed on the background class, only).

We aim to improve segmentation performance by augmenting the background class with auxiliary context classes. Specifically, we propose to utilize context labels assigned to different and decomposed background regions. In order to divide the background class into $t>1$ classes, we create another model $\tilde{f}_{\theta}(\cdot)$ with $\tilde{\boldsymbol{z}}_i = \tilde{f}_{\theta}(\boldsymbol{x}_i) \in \Reals^{c+t-1}$ and predicted probability $\tilde{\boldsymbol{p}}_i$. We also require an additional one-hot label $\tilde{\boldsymbol{y}}_{i} \in \Reals^{c+t-1}$, where we require $\tilde{y}_{ij} = y_{ij}$ $\forall$ $i$, $j \in \{1, \cdots, c-1\}$ and $\sum_{j=c}^{c+t-1}\tilde{y}_{ij} = y_{ic}$ $\forall$ $i$. With this notion, the segmentation loss can be written as:

\begin{equation}
    \begin{split}
    & \mathcal{L}_{seg}(\tilde{f}_\theta(\boldsymbol{x}_i),\tilde{\boldsymbol{y}}_{i}) = \sum_{j=1}^{c+t-1} \mathcal{L}(\tilde{p}_{ij}, \tilde{y}_{ij}) \\
    & = \underbrace{\sum_{j=1}^{c-1}\mathcal{L}(\tilde{p}_{ij}, \tilde{y}_{ij})}_{\text{ROI classes}} + \underbrace{\sum_{j=c}^{c+t-1}\mathcal{L}(\tilde{p}_{ij}, \tilde{y}_{ij})}_{\text{background classes}},
    \label{eq:context}
    \end{split}
\end{equation}

In the following method section, we always consider a simplified but common case where we only have one ROI class ($c=2$) for simplicity. In other words, we only consider binary ROI segmentation with $f_\phi(\boldsymbol{x}_i) \in \Reals^2$, although the method can be naturally extended to multi-class ROI segmentation. With this assumption, Eq.~\ref{eq:context} can be simplified as:

\begin{equation}
    \mathcal{L}_{seg}(\tilde{f}_\theta(\boldsymbol{x}_i),\tilde{\boldsymbol{y}}_{i}) = \underbrace{\mathcal{L}(\tilde{p}_{i1}, \tilde{y}_{i1})}_{\text{ROI class}} + \underbrace{\sum_{j=2}^{t+1}\mathcal{L}(\tilde{p}_{ij}, \tilde{y}_{ij})}_{\text{background classes}}
    \label{eq:simplifiedcontext}
\end{equation}

\subsection{False positives on background class}

To better understand the effect of the background class on the model learning, we train multiple CNNs on segmentation datasets which contain heterogeneous background. We conduct experiments on challenging tasks including liver tumor segmentation in computed tomography (CT) images~\cite{bilic2019liver}, kidney tumor segmentation in CT images~\cite{heller2019kits19}, colon tumor segmentation in CT images~\cite{simpson2019large}, vestibular schwannoma (VS) segmentation in T2-weighted magnetic resonance (MR) images~\cite{shapey2019artificial}, brain stroke lesion segmentation in T1-weighted MR images~\cite{liew2018large} and pancreatic tumor segmentation in CT images~\cite{simpson2019large}. We adopt a well-configured 3D U-Net~\cite{isensee2021nnu} as the segmentation model for all the experiments, which has been demonstrated to yield competitive results across different medical image segmentation tasks. The detailed data and network configurations are summarized in Section~\ref{sec:experiments}.

Visualizations of the segmentation results are shown in Fig.~\ref{fig_segexample}(a). When trained with binary segmentation tasks and without context labels, the models are prone to over-segment the ROIs with many FP. Specifically, the model trained on the binary liver tumor task predicts other organs outside liver as liver tumor; the model trained on the binary kidney tumor task predicts parts of the healthy kidney regions as kidney tumor; while the model trained on the binary brain tumor task predicts the surrounding brain tissue as brain tumor; the model trained on the binary brain lesion task predicts unrelated healthy brain regions as brain lesion.

\begin{figure*}[t]
\centering
\includegraphics[width=\textwidth]{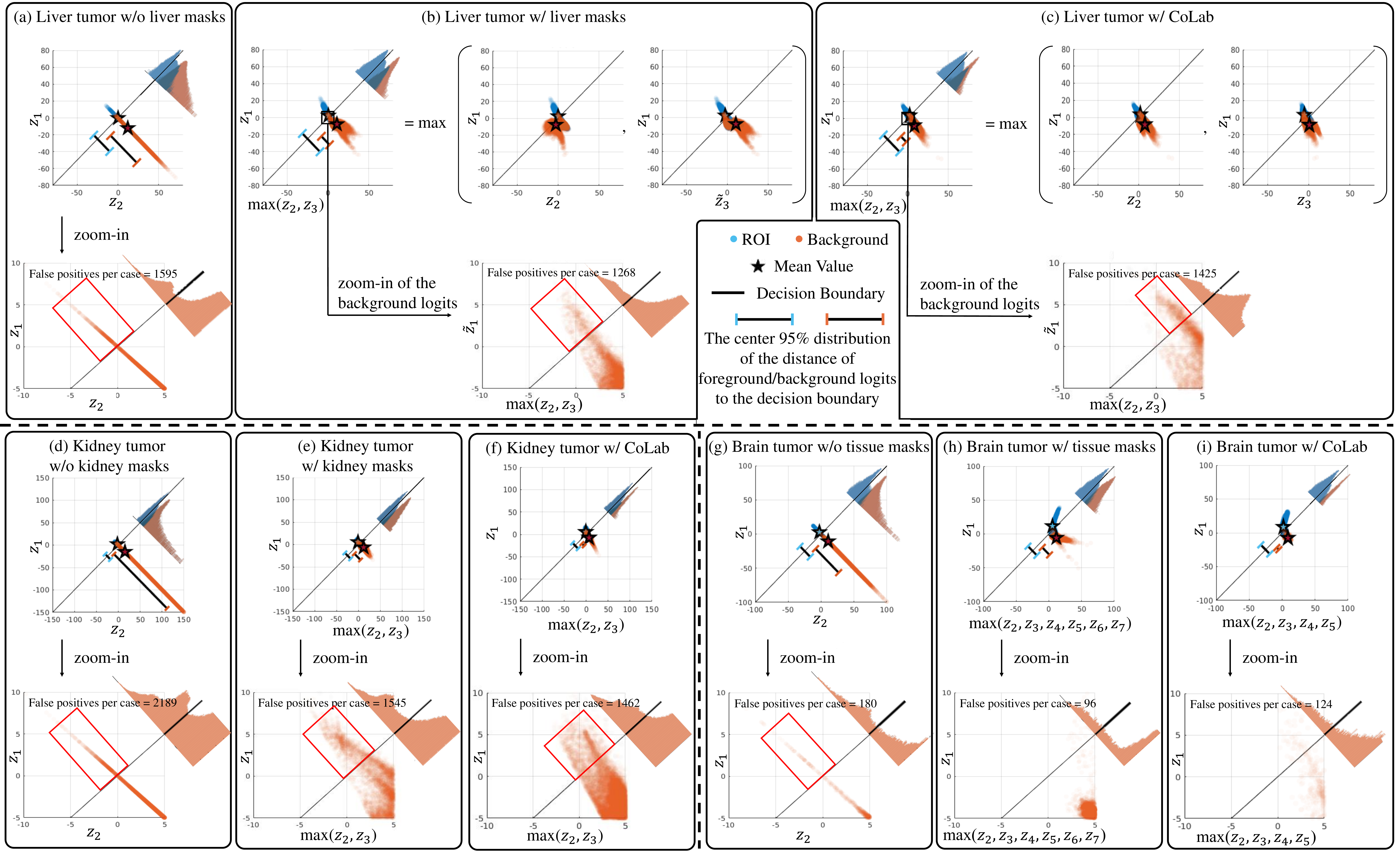}

\caption{The distribution of activations (logits) after the classification layer using ROI (blue) or background samples (orange) of test data as network input. Here, $z_1$ denotes logits for ROI classes, while $z_i \;  (i>1)$ denotes logits for background classes.  We also report the number of pixel-level predictions that are FP. Without context labels (a, d, g), CNNs fail to map the background samples to compact clusters. This is because CNNs are forced to learn generic filters for the background samples, yielding similar activations regardless of different background structures with diverse patterns/appearance. Consequently, the logit distribution of background samples would spread across the decision boundary, leading to increased FP. Extending the label space with anatomy masks (b, e, h) or CoLab (c, f, i) can alleviate this issue.} \label{fig_logitmap}
\end{figure*}

\begin{figure}[t]
\centering
\includegraphics[width=0.48\textwidth]{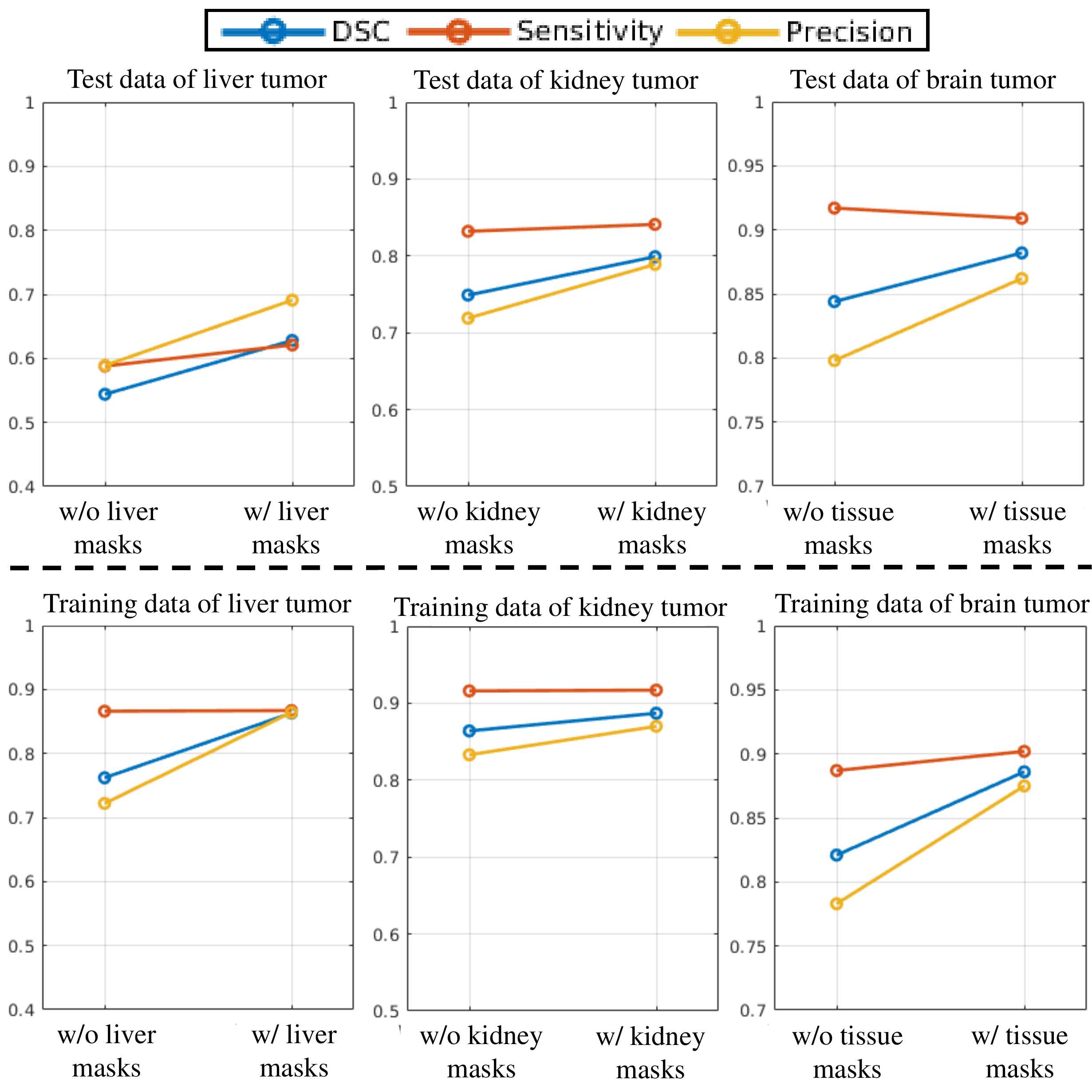}

\caption{Performance on liver tumor, kidney tumor and brain tumor segmentation on training data (lower part) and testing data (upper part) when networks were trained w/o and w/ context labels. Extending the label space with anatomy masks can help the model improve the performance by increasing the precision, while the sensitivity is largely retained.} \label{fig_segperformance}
\end{figure}

\subsection{Underfitting of background samples}

To study the model behavior when trained with heterogeneous background classes, we can monitor the logit distribution of samples from ROI and background classes for the test data. Our observations for liver tumor, kidney tumor, and brain tumor segmentation are summarized in Fig.~\ref{fig_logitmap}(a, d, g). 

We find that the CNN models map the ROI samples to a compact cluster in the logit space while background samples form a more dispersed distribution. This indicates that the model cannot easily map all background samples to a single cluster representing the background class. Although the model seems to separate ROI from background samples in the feature space, it builds complex background representations and unable to capture an accurate decision boundary between ROI and immediate context. A possible reason is that the CNN uses most of its capacity to extracted the common features among background regions with different characteristics. Unfortunately, these shared features are not very discriminate. Specifically, we observe that the logit distribution of background samples overlaps with the learned decision boundary. This is the reason why the model predicts many FP leading to over-segmentation of ROI structures when the background class is heterogeneous. We hypothesise that the width of the distribution serves as an indicator of the heterogeneity of a specific class and is a sign of the difficulty during learning. As a result of sample heterogeneity, a CNN may struggle to reduce intra-class variation of the background class and underfit the background samples, failing to recognize the background samples that share similar characteristics with ROI samples. 

It should be noted that we do not observe significant difference between the logit distribution of training background samples and test background samples, as also shown in previous studies~\cite{li2019overfitting, li2020analyzing}, indicating that the logit shift of background samples is indeed due to underfitting instead of overfitting.

\subsection{The effect of context labels}

The availability of context labels greatly helps with the ROI segmentation, which provide additional signals to CNN to fit the training data of heterogeneous background samples. We confirm this empirically by including human-defined context labels in the model training. Specifically, we adopt liver masks ($t=2$) for liver tumor segmentation, kidney masks ($t=2$) for kidney tumor segmentation, brain tissue masks including ventricles, deep grey matter, cortical grey matter, white matter and other tissues ($t=6$) for brain tumor segmentation. The kidney and liver masks are manually annotated, while the brain tissue masks are generated automatically using the paired T1-weighted MR images and a multi-atlas label propagation with expectation–maximisation based refinement (MALP-EM)~\cite{ledig2015robust}. In order to make a fair comparison, we make sure that all the experiments share the same training schedule except the label space. We sample the training patches only considering ROIs. Specifically, we make sure 50\% training patches to contain ROIs and sample the other half of the training patches uniformly.

The observations on test and training set are summarized in Fig.~\ref{fig_segperformance}. We find that the models yield overall better performance when trained with anatomy masks, as indicated by improved DSC (defined as $ DSC=2\frac{\text{sensitivity} \cdot \text{precision}}{\text{sensitivity} + \text{precision}}$). Furthermore, we observe that the model trained with context labels yields higher precision while preserving similar sensitivity. The observation is consistent with the visualization results in Fig.~\ref{fig_segexample}(b), where we find the models trained with human-defined context labels reduce FP.

Similarly, we visualize the corresponding network behaviour in Fig.~\ref{fig_logitmap}(b, e, h). As we only consider the segmentation performance of the ROI (class 1), we visualize the logits in the plane of ($z_1$, max($z_2$, ..., $z_{t+1}$)). We observe that the model trained with anatomy masks maps the background samples to a narrower distribution and reduces the background logits shift across the decision boundary. This indicates that the models fit the training data better with the help of context labels. Instead of building generic filters for all the background samples, the CNNs can dedicate specific filters to model a more homogeneous subparts of the background samples that share common characteristics. The models are faced with a simplified segmentation task with homogeneous background subclasses yielding better overall performance with the same model capacity.

Although anatomy masks are found to be effective context labels, they are not always available in real-world applications. Specifically acquiring manually annotated context labels is time-consuming and would require significant efforts from human experts to generate annotations at large scale. Therefore, we propose CoLab which can automatically discover specific soft context labels using a meta-learning strategy. CoLab benefits the segmentation model training by making it to fit the background samples better, achieving comparable or even better performance when compared with models trained on manually defined anatomy masks.

\section{CoLab}
\label{sec:method}

\subsection{Overview}
\label{sec:method-overview}

\begin{figure*}[t]
\centering
\includegraphics[width=\textwidth]{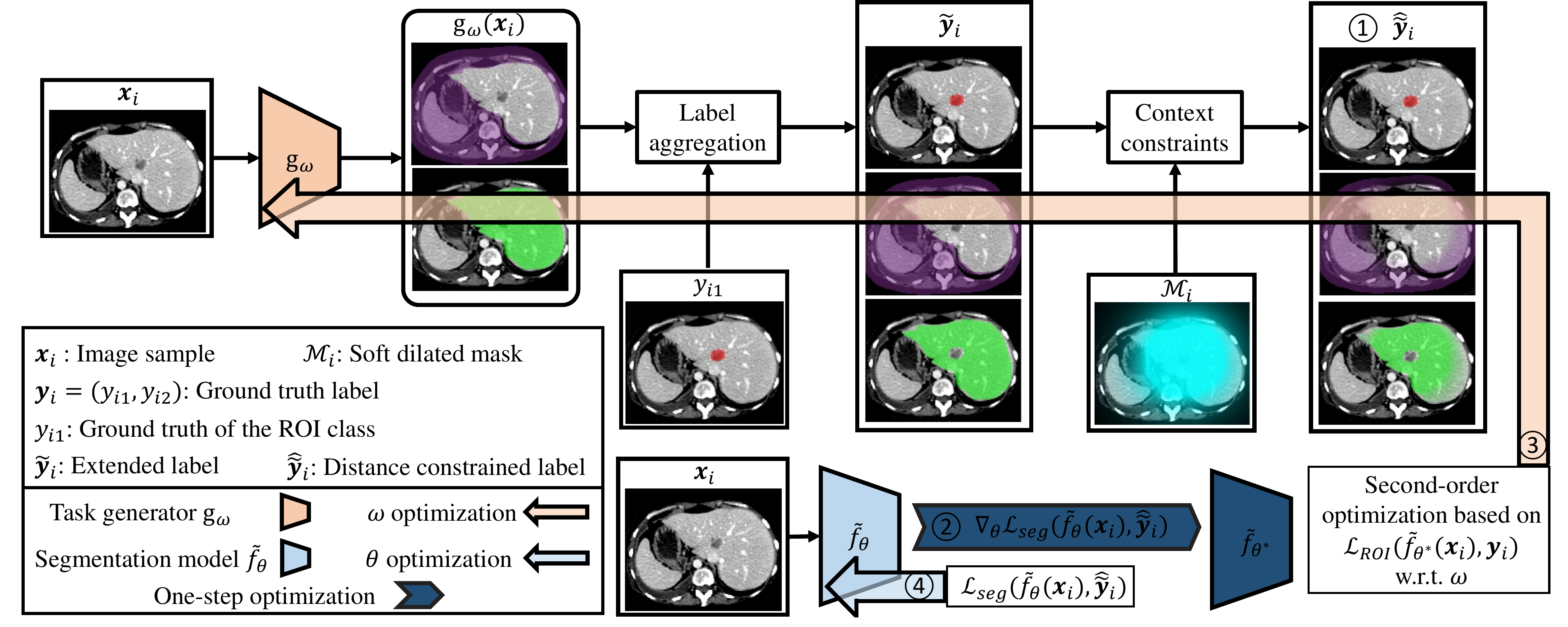}

\caption{Illustration of the training process of CoLab. $g_\omega(\cdot)$ is optimized to generate context labels such that it can help reduce the training loss of the ROI in the most effective way, see Section \ref{sec:method-overview} for more details. } \label{fig_method}
\end{figure*}

Now we consider CNNs for binary semantic segmentation. Typically, we are given a baseline segmentation model $f_\phi(\cdot)$ that maps the input image $\boldsymbol{x}_i$ to the label space $f_\phi(\boldsymbol{x}_i) \in \Reals^2$. In order to fit the label space with context labels, we first extend the classification layer of $f_\phi(\cdot)$ with additional $t-1$ output neurons and obtain $\tilde{f}_{\theta}(\cdot)$ which map $\boldsymbol{x}_i$ to $\tilde{f}_{\theta}(\boldsymbol{x}_i) \in \Reals^{t+1}$. We employ another model $g_\omega(\cdot)$ as the task generator parameterized by $\omega$ to produce context labels, with the network output $\boldsymbol{o}_i$ = $g_\omega(\boldsymbol{x}_i) \in \Reals^t$. We have no requirements for the backbone of  $g_\omega(\cdot)$ and empirically keep it the same with that of $f_\phi(\cdot)$. 

We illustrate the training process of the proposed CoLab in Fig.~\ref{fig_method}. At the beginning of each iteration, we obtain the context labels termed as distance constrained label $\hat{\tilde{\boldsymbol{y}}}_i$ by taking use of $\boldsymbol{o}_i$ and the ground truth $\boldsymbol{y}$. This process is marked with \textcircled{1} and will be illustrated in Section~\ref{sec:method-labelagg} and~\ref{sec:method-contcons}. Then, we calculate a new $\theta^*$ with one step of gradient descent with \textcircled{2} to access the impact of $\hat{\tilde{\boldsymbol{y}}}_i$ on model training of ROI segmentation. Next, we optimize $\omega$ based on the second-order derivatives through a meta learning scheme with \textcircled{3}, which would be demonstrated in Section~\ref{sec:meta-learning}. Finally, we optimize the segmentation model $\tilde{f}_{\theta}(\cdot)$ based on the updated context labels with \textcircled{4}.

\subsection{Label aggregation}
\label{sec:method-labelagg}

We calculate the context probability $(q_{ij})_{j=1}^{t}$ based on $\boldsymbol{o}_i$ via the softmax function as:

\begin{equation}
    q_{ij} = \frac{\mathrm{e}^{o_{ij}}}{\sum_{j=1}^{t}\mathrm{e}^{o_{ij}}}.
    \label{eq:generatedclass}
\end{equation}

The extended label $\tilde{\boldsymbol{y}}_i \in \Reals^{t+1}$ is calculated by aggregating the original label $\boldsymbol{y}_i = (y_{i1}, y_{i2})$ and the context probability $\boldsymbol{q}_i$ with:

\begin{equation}
\begin{aligned}
\tilde{y}_{ij}=\left\{
\begin{array}{rcl}
y_{i1}         &      & \text{if} \quad j = 1,\\
q_{ij}         &      & \text{if} \quad j > 1 \quad \text{and} \quad y_{i1}=0,\\
0       &   & \text{otherwise.}
\end{array} \right.
\label{eq:aggfunc}
\end{aligned}
\end{equation}

\begin{figure}[t]
\centering
\includegraphics[width=0.48\textwidth]{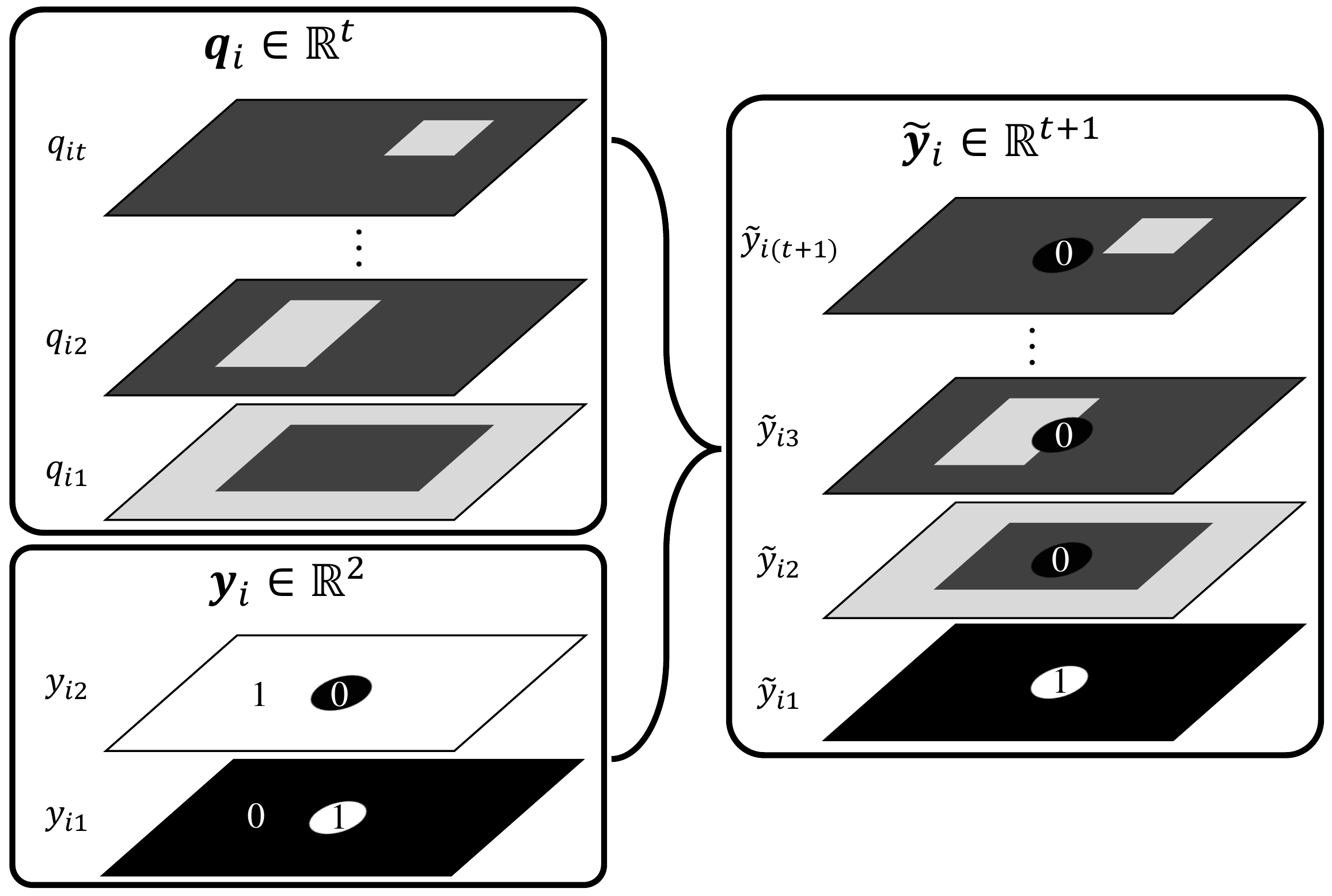}

\caption{Illustration of the label aggregation process. We generate the extended label $\tilde{\boldsymbol{y}}_{i}$ by aggregating the original label $\boldsymbol{y}_i$ with the context probability $\boldsymbol{q}_i$.} \label{fig_labelagg}
\end{figure}

The label aggregation process is also illustrated in Fig.~\ref{fig_labelagg}. By doing so, we can decompose the background class $y_{i2}$ into $t$ subclasses while ensure that the $\tilde{\boldsymbol{y}}_i$ contains sufficient information about the ROI segmentation. For multi-class segmentation with totally $c$ classes, $\tilde{\boldsymbol{y}}_i$ can be calculated as:

\begin{equation}
\begin{aligned}
\tilde{y}_{ij}=\left\{
\begin{array}{rcl}
y_{ij}         &      & \text{if} \quad j < c,\\
q_{ij}         &      & \text{if} \quad j \geq c  \quad \text{and} \quad y_{ij}=0 \quad \forall j<c, \\
0       &   & \text{otherwise.}
\end{array} \right.
\label{eq:aggfunc2}
\end{aligned}
\end{equation}

\subsection{Context constraints}
\label{sec:method-contcons}

Compared with background samples which are further away, the ones closer to the ROIs share similar characteristics with ROI samples and are more likely to be misclassified. In order to make the segmentation model focus more on those hard background samples that are close to ROIs, we make an assumption that all samples which are distant from the ROI need less attention and should be safely assigned the same background label. Specifically, we create a hard label $\boldsymbol{b}_i \in \Reals^{t+1}$ to represent the the background samples that are far from the ROI:

\begin{equation}
b_{ij}=\left\{
\begin{array}{rcl}
1         &      & \text{if} \quad j = 2,\\
0       &      & \text{otherwise}.
\end{array} \right.
\label{eq:bglabel}
\end{equation}

By utilizing the ROI label $\boldsymbol{y}_i$, we calculate the corresponding distance map $d_i$ which is the Euclidean distance of a pixel to the closest boundary point of the ROI of any class. We set $d_i$ to be zero for pixels inside the ROI.

\begin{figure}[t]
\centering
\includegraphics[width=0.48\textwidth]{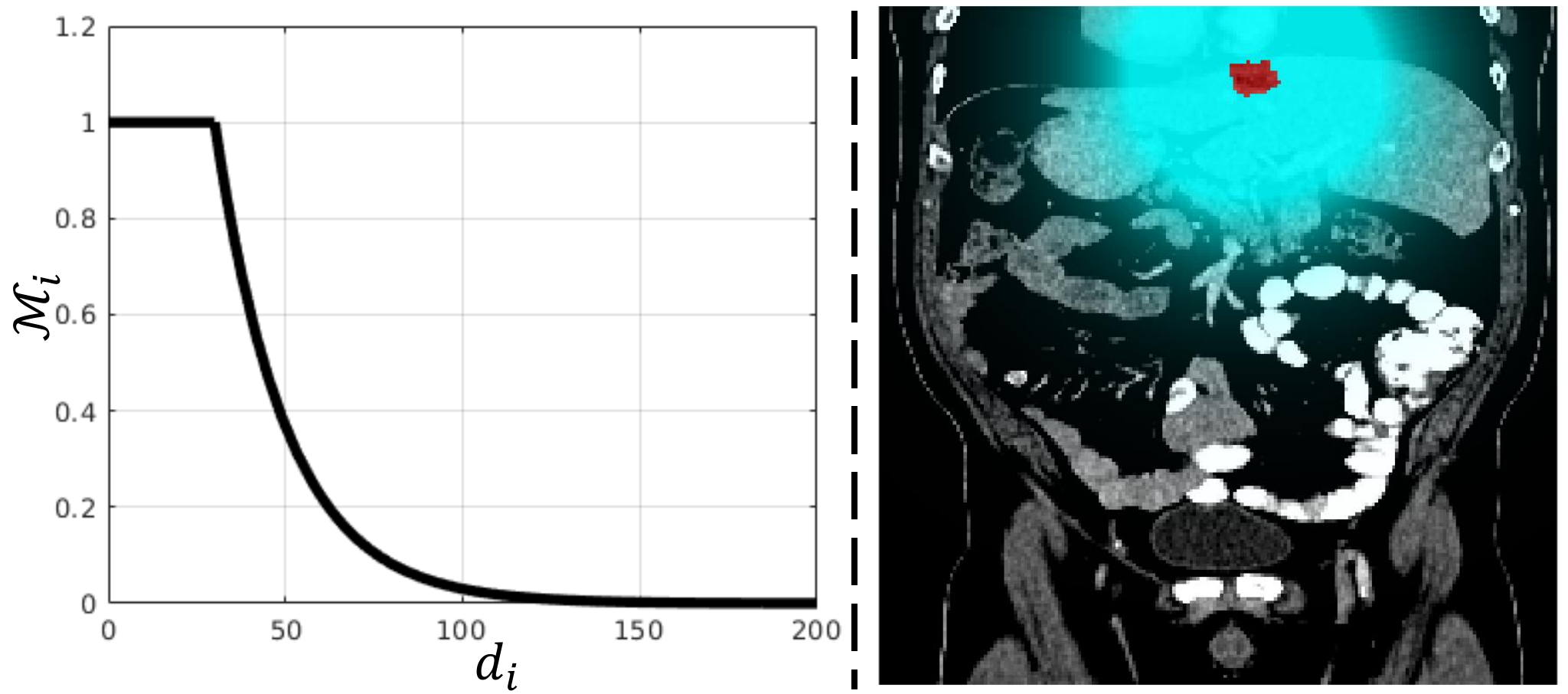}

\caption{The task generator focuses on the context which is near to the ROI. To do so, we train the segmentation model to classify the pixels which are outside the soft dilated ROI as the same class (pixels marked without color in the right figure). (Left part) The value of a soft dilated mask $\mathcal{M}_i$ with $m=30$ and $\tau=20$. (Right part) The visualization of the dilated mask $\mathcal{M}_i$ (blue) with a liver tumor (red) of a training case.} \label{fig_lossmask}
\end{figure}

We then calculate a soft dilated mask $\mathcal{M}_i$ based on $d_i$:

\begin{equation}
    \mathcal{M}_i =\left\{
    \begin{array}{rcl}
     1         &      & \text{if} \quad d_i < m,\\
     \mathrm{e}^{\frac{-d_{i}+m}{\tau}}       &      & \text{otherwise,}
    \end{array} \right.
    \label{eq:lossmask}
\end{equation}

where $m$ is the margin controlling the model's focus on the pixels which are neighbouring to ROI and $\tau$ is the temperature to control the probabilities of the dilated regions. Empirically, $m$ is set as 30 and $\tau$ is set as 20 for all our experiments. $\mathcal{M}_i$ would be 1 for pixels around ROIs and decrease close to 0 for pixels far from ROIs. We visualize an example of $\mathcal{M}_i$ with liver tumor in Fig.~\ref{fig_lossmask}. The distance constrained label then is calculated as:

\begin{equation}
    \hat{\tilde{\boldsymbol{y}}}_i=\mathcal{M}_i\tilde{\boldsymbol{y}}_i+(1-\mathcal{M}_i)\boldsymbol{b}_i.
    \label{eq:masky}
\end{equation}

In this way, only the regions neighbouring to the tumor are considered to be classified as the contextual background class. Specifically, both $\tilde{f}_{\theta}(\cdot)$ and $g_\omega(\cdot)$ would be trained to focus on the regions which are close to ROIs.

\subsection{Task generator optimization with meta-gradients}
\label{sec:meta-learning}

We formulate the optimization of CoLab as a bi-level problem:

\begin{align}
    &\min_{\omega}\frac{1}{N}\sum_{i=1}^{N} \mathcal{L}_{ROI}(\tilde{f}_{\theta^*}(\boldsymbol{x}_i), \boldsymbol{y}_i) \label{eqn:objectivefunA} \\
    &s.t.\quad \theta^* = \argmin_{\theta}\frac{1}{N} \sum_{i=1}^{N}\mathcal{L}_{seg}(\tilde{f}_{\theta}(\boldsymbol{x}_i), \hat{\tilde{\boldsymbol{y}}}_i)),
    \label{eq:objectivefunB}
\end{align}

where $\mathcal{L}_{ROI}(\tilde{f}_{\theta}(\boldsymbol{x}_i), \boldsymbol{y}_i) = \mathcal{L}_c(\tilde{p}_{i1}, y_{i1})$ is computed on the ROI class. Different from $\mathcal{L}$, we choose $\mathcal{L}_c$ to be a criteria that can be calculated in the one-versus-all manner, such as binary cross entropy (BCE) and soft DSC loss, to represent the binary segmentation performance of the ROI class with more than two output logits.

We train the model using a batch of training samples with batch size $n$. For simplicity, we shorten $\frac{1}{n}\sum_{i=1}^{n} \mathcal{L}_{ROI}(\tilde{f}_{\theta^*}(\boldsymbol{x}_i), \boldsymbol{y}_i)$ as $\mathcal{L}_{ROI}(\theta^*)$ and $\frac{1}{n}\sum_{i=1}^{n}\mathcal{L}_{seg}(\tilde{f}_{\theta}(\boldsymbol{x}_i), \hat{\tilde{\boldsymbol{y}}}_i)$ as $\mathcal{L}_{seg}(\theta, \omega)$ in the following descriptions. The bi-level optimization problem defined in Eq.~\ref{eqn:objectivefunA} and \ref{eq:objectivefunB} can be solved with gradient descent~\cite{pedregosa2016hyperparameter, franceschi2018bilevel}. Specifically, the derivative of $\mathcal{L}_{ROI}(\theta^*)$ w.r.t. $\omega$ can be calculated by applying the chain rule: 

\begin{equation}
    \nabla_{\omega}\mathcal{L}_{ROI}(\theta^*) = (\frac{\partial \theta^*}{\partial \omega})^\intercal \nabla_{\theta}\mathcal{L}_{ROI}(\theta^*),
    \label{eq:funcchain}
\end{equation}

where $\nabla_{\omega} = (\frac{\partial}{\partial \omega})^\intercal$ and $\nabla_{\theta} = (\frac{\partial}{\partial \theta})^\intercal$. One can compute $\frac{\partial \theta^*}{\partial \omega}$ based on implicit function theorem~\cite{bengio2000gradient}. However, the derived result would contain a Hessian which is computational expensive and not always possible to access for deep neural networks. Among the many heuristics for the gradient approximation~\cite{pedregosa2016hyperparameter, franceschi2018bilevel, liu2018darts}, we follow the solution described in~\cite{finn2017model, liu2018darts} to approximate $\theta^*$ by a single optimization step. Specifically, we sample a batch of training data and approximate the optimal inner variable $\theta^*$ with a step of gradient decent:

\begin{equation}
    \theta^* \approx  \theta-\alpha \nabla_\theta  \mathcal{L}_{seg}(\theta, \omega),
    \label{eq:innerupdate}
\end{equation}

where $\alpha$ is step size, which is kept the same with the learning rate of $\theta$. We differentiate this equation w.r.t. $\omega$ from both sides yielding:

\begin{equation}
    \frac{\partial \theta^*}{\partial \omega} = -\alpha\nabla^2_{\theta,\omega}\mathcal{L}_{seg}(\theta^*, \omega),
    \label{eq:omegasimple}
\end{equation}

where $\nabla^2_{\theta,\omega} = \frac{\partial \nabla_{\theta}}{ \partial \omega}$. By substituting Eq.~\ref{eq:omegasimple} into Eq.~\ref{eq:funcchain}, we can obtain the gradient on $\omega$ and update it with:

\begin{equation}
\begin{aligned}
    & \omega^{t+1} = \omega^t - \beta \nabla_{\omega^t}\mathcal{L}_{ROI}(\theta^*) \\
    & = \omega^t + \alpha\beta\nabla^2_{\omega^t,\theta}\mathcal{L}_{seg}(\theta^*, \omega^t) \nabla_{\theta}\mathcal{L}_{ROI}(\theta^*),
    \label{eq:func3s}
\end{aligned}
\end{equation}

where $\beta$ is the learning rate to update $\omega$. In this way, the task generator $g_{\omega}$ is explicitly trained to produce effective context labels with the second-order gradients. Although Eq.~\ref{eq:func3s} contains an expensive vector-matrix product, it is feasible to calculate with prevailing machine learning frameworks such as PyTorch~\cite{paszke2019pytorch}. We find CoLab can be efficient as it costs as little as 30\% additional training time. We summarize the implementation details in supplementary material. The full algorithm is summarized in Algorithm~\ref{alg:CoLab}.

\begin{algorithm}
\caption{Context label learning (CoLab)}
\label{alg:CoLab}
\begin{algorithmic}[1]
\Require
      \Statex $\mathcal{D} = \{ (\boldsymbol{x}_i, \boldsymbol{y}_i) \}_{i = 1}^N$: training data; $t$: the number of context classes, $f_\phi(\cdot)$: the segmentation model which produces $f_\phi(\boldsymbol{x}_i) \in \Reals^2$; $g_\omega(\cdot)$: the task generator which produce $g_\omega(\boldsymbol{x}_i) \in \Reals^t$.
      \Statex $\alpha$, $\beta$: learning rates to update $\theta$ and $\omega$.
      
\State Extend the classification layer of $f_\phi(\cdot)$ to get $\tilde{f}_{\theta}(\cdot)$ which produce $\tilde{f}_{\theta}(\boldsymbol{x}_i) \in \Reals^{t+1}$.

\For{each iteration} 

\State Sample a batch of data $\mathcal{B}=\{ (\boldsymbol{x}_i, \boldsymbol{y}_i) \}_{i = 1}^n$ from $\mathcal{D}$.

\For{a number of steps} \Comment{\emph{\small{Note: One step is sufficient in our experiments.}}}

\State Generate the context probability $\boldsymbol{q}_i$ based on the output of task generator $g_\omega(\boldsymbol{x}_i)$ via Eq.~\ref{eq:generatedclass}.

\State Generate the extended label $\tilde{\boldsymbol{y}}_{i}$ by aggregating the context probability $\boldsymbol{q}_i$ with $\boldsymbol{y}_i$ via Eq.~\ref{eq:aggfunc}.

\State Calculate a loss mask $\mathcal{M}_i$ via Eq.~\ref{eq:lossmask} and generate the distance constrained label $\hat{\tilde{\boldsymbol{y}}}_i$ via Eq.~\ref{eq:masky}.

\State Calculate a new $\theta^*$ with the gradient descent optimization algorithm via Eq.~\ref{eq:innerupdate}.

\State Optimize $\omega$ based on meta-gradients via Eq.~\ref{eq:func3s}. \Comment{\emph{\small{Training the task generator.} }}

\EndFor

\State Update $\theta$ with gradient decent based on updated $\omega$. \Comment{\emph{\small{Training the segmentation model.} }}

\EndFor
\end{algorithmic}
\end{algorithm}

\section{Experiments}
\label{sec:experiments}

\subsection{Experimental setup}

\subsubsection{Network configurations}

We use a state-of-the-art 3D U-Net~\cite{isensee2021nnu} as the network backbone for both the segmentation model $f_\phi$/$\tilde{f}_{\theta}$ and the task generator $g_\omega$. We normalize all datasets with the built-in pipeline following~\cite{isensee2021nnu}. Specifically, we adopt case-wise Z-score normalization within brain masks for MR images, while we employ dataset-wise Z-score normalization based on ROI samples for CT images after clipping the Hounsfield units (HU) from 0.5\% to 99.5\%. We utilize the default data augmentation policies for all experiments. We adopt a combination of CE and sample-wise soft DSC loss with equal weight for $\mathcal{L}$ while using BCE for $\mathcal{L}_c$. We use a batch size of 2 and patch size of 80$\times$80$\times$80. We train the networks for 1000 epochs for brain tumor and lesion segmentation and 2000 epochs for liver, kidney, colon and pancreas tumor segmentation  as we observed that the network needed more iterations to converge latter tasks. We summarize the hyper-parameters of all experiments in supplementary material. All reported results are the average of two runs with different random seeds.

\subsubsection{Liver tumor segmentation} 

We evaluate CoLab for liver tumor segmentation with the training dataset from the Liver Tumor Segmentation Challenge (LiTS) which contains 131 CT images. We exclude the samples which do not contain any liver tumor leaving 118 cases. We resample all CT images to a common voxel spacing of 1.9$\times$1.9$\times$2.5 mm following~\cite{isensee2021nnu}. We train models with 83 cases and test on 35 cases.

\subsubsection{Kidney tumor segmentation}

We further conduct experiments using the training dataset of the Kidney Tumor Segmentation Challenge (KiTS) which contains 210 CT images. We resample all CT images to a common voxel spacing of 1.6$\times$1.6$\times$3.2 mm following~\cite{isensee2021nnu}. We tested on 70 cases and used the other 140 cases as the training data.

\subsubsection{Colon tumor segmentation}

We evaluate CoLab for the case of colon tumor segmentation from CT images. We collect 126 colon cancer CT images from the training dataset of the Medical Segmentation Decathlon challenge~\cite{simpson2019large}. We resample all CT images to the voxel spacing of 1.6$\times$1.6$\times$3.1 mm following~\cite{isensee2021nnu}. We train models with 88 cases and test on the other 38 cases.

\subsubsection{Brain tumor segmentation}

We also conduct experiments for brain tumor segmentation using the VS dataset~\cite{shapey2019artificial} which contains 243 paired T1-weighted and T2-weighted MR images. We only used T2-weighted MR images for evaluating CoLab but used T1-weighted MR images to generate the brain structure masks with MALP-EM. We did not use brain masks for the histogram normalization of this task because VS can appear at the brain boundary and may be excluded when using common brain extraction algorithms. All MR images have the same isotropic voxel spacing of 1.0 mm$^3$. We train the models with 176 cases and test on 46 cases following~\cite{shapey2019artificial}.

\subsubsection{Brain stroke lesion segmentation}

Additionally, we evaluate CoLab with brain stroke lesion segmentation using the Anatomical Tracings of Lesions After Stroke (ATLAS) dataset~\cite{liew2018large} which contains 220 T1-weighted MR images. The MR images all have the same voxel spacing of 1.0 mm$^3$. We randomly selected 145 cases as training data and left the rest 75 cases for testing.

\subsubsection{Pancreas and pancreatic tumor mass segmentation}

We further evaluate Colab in the setting of multi-class segmentation. Specifically, we train the segmentation model to segment three classes including pancreas, pancreatic tumor mass and background. We aim to find the context labels which can benefit the segmentation of both foreground classes. We collect 281 CT images containing pancreas tumor from the training dataset of the Medical Segmentation Decathlon challenge~\cite{simpson2019large, attiyeh2018survival, attiyeh2019preoperative, chakraborty2018ct}. We resample all the CT images to the voxel spacing of 1.3$\times$1.3$\times$2.6 mm following~\cite{isensee2021nnu}. We randomly split the dataset into 197 cases for training and 84 cases for testing.

\subsection{Compared methods and processing}

\subsubsection{Context labels based on k-means}

We compare CoLab with alternative approaches for context label generation, including a context label generation via clustering. Here, we take pixels inside the body masks or brain masks as samples and employ $k$-means~\cite{arthur2006k} to construct $t$ clusters. We kept $t$ the same with the class number of human-defined anatomy masks. Specifically, we chose $t=2$ for liver, kidney, colon and pancreas tumor segmentation and $t=6$ for brain tumor and brain stroke lesion segmentation.

\subsubsection{Context labels based on dilated masks}

We also compare with a baseline using dilated masks of the ROI as the context labels. This idea is somewhat similar to label smoothing~\cite{muller2019does,islam2021spatially} where models are trained with blurred ROI labels. Specifically, we take the soft dilated masks $\mathcal{M}_i$ defined in Eq.~\ref{eq:lossmask} and set the context probability as $\boldsymbol{p}_i=[1-\mathcal{M}_i, \mathcal{M}_i]^\intercal$.

\subsubsection{Context labels predicted with external datasets}

We further evaluate and compare to an approach which leverages prior knowledge from other datasets. Specifically, we trained a segmentation model with 20 CT images using data from~\cite{xu2015efficient} which contains labels of 14 abdominal organs including liver and kidney. We resample the 20 CT images to the voxel spacing of 1.6$\times$1.6$\times$3.2 mm. We reduce the dependency of the model on longitudinal axis and trained the segmentation model with a patch size of 128$\times$128$\times$32 as the slice numbers differ across datasets. We then apply this segmentation model to the resampled training split of LiTS as well as KiTS, and extract the liver and kidney masks as the automatically generated contextual anatomy masks.

\subsubsection{Post-processing}

We also compare with a common strategy to suppress FP in segmentation based on component-based post-processing, which is widely adopted in many segmentation pipelines~\cite{isensee2021nnu}. Specifically, we assume there is always one ROI and remove all but the largest region. For the cases when ROIs contain multiple classes (pancreas and pancreatic tumor segmentation), we take all the ROIs as a whole and only keep the largest component.

\subsection{Quantitative results}

\begin{table*}[t]
\centering
\caption{Evaluation of different binary segmentation tasks with different types of context label generation approaches. We divide the results according to whether expert knowledge and labeling efforts are included. The best and second best results without human efforts are in \textbf{bold} with the best also \underline{\textbf{underlined}}.}\label{tabresults}
\newsavebox{\tablekits}
\begin{lrbox}{\tablekits}

\begin{tabular}{m{30mm}<{\centering}|m{50mm}<{\centering}|m{5mm}<{\centering}|m{10mm}<{\centering}|m{10mm}<{\centering}|m{10mm}<{\centering}|m{10mm}<{\centering}}
\hlineB{3}
Task & Method & $t$ & DSC & SEN & PRC & HD
\\
\hlineB{1}
\multirow{8}{*}{\tabincell{c}{Liver tumor~\cite{bilic2019liver}}}& w/o liver masks & 1 & 54.4 & 58.8 & 58.9 & 111.1 \\
& $K$-means~\cite{arthur2006k} & 2 & \textbf{61.4} & \textbf{61.4} & 67.0 & 71.9 \\
& Dilated masks~\cite{muller2019does} & 2 & 60.7 & 59.8 & \textbf{\underline{68.0}} & 67.6 \\
& CoLab & 2 & \textbf{\underline{62.5}} & \textbf{\underline{62.8}} & \textbf{67.3} & 69.4 \\
& CoLab & 4 & 57.3 & 60.5 & 62.3 & \textbf{56.3} \\
& CoLab & 6 & 59.7 & 60.3 & 65.2 & \textbf{\underline{43.6}} \\
\cline{2-7}
& w/ model-predicted liver masks~\cite{xu2015efficient} & 2 & 62.4 & 61.6 & 70.6 & 44.1 \\
& w/ liver masks~\cite{bilic2019liver} & 2 & 62.8 & 62.1 & 69.1 & 53.5 \\
\hline
\multirow{8}{*}{\tabincell{c}{Kidney tumor~\cite{heller2019kits19}}}& w/o kidney masks & 1 & 74.9 & 83.2 & 71.9 & 120.4 \\
& $K$-means~\cite{arthur2006k} & 2 & \textbf{76.8} & \textbf{83.5} & 74.3 & 87.1 \\
& Dilated masks~\cite{muller2019does} & 2 & 76.4 & \textbf{\underline{83.9}} & 73.1 & 95.3 \\
& CoLab & 2 & \textbf{\underline{78.5}} & 82.2 & \textbf{\underline{77.7}} & \textbf{75.7} \\
& CoLab & 4 & 76.4 & 80.6 & \textbf{76.5} & \textbf{\underline{63.7}} \\
& CoLab & 6 & 74.9 & 81.0 & 73.3 & 79.4 \\
\cline{2-7}
& w/ model-predicted kidney masks~\cite{xu2015efficient} & 2 & 79.2 & 81.3 & 82.7 & 38.1 \\
& w/ kidney masks~\cite{heller2019kits19} & 2 & 79.9 & 84.1 & 78.9 & 54.7 \\
\hline
\multirow{6}{*}{\tabincell{c}{Colon tumor~\cite{simpson2019large}}}& w/o context labels & 1 & 45.6 & \textbf{\underline{53.9}} & 45.3 & 154.9 \\
& $K$-means~\cite{arthur2006k} & 2 & 44.9 & 47.2 & 48.7 & 111.7 \\
& Dilated masks~\cite{muller2019does} & 2 & \textbf{47.1} & 49.8 & \textbf{\underline{52.8}} & \textbf{83.3} \\
& CoLab & 2 & \textbf{\underline{48.9}} & \textbf{\underline{53.9}} & 50.3 & 86.0 \\
& CoLab & 4 & 45.3 & 48.1 & \textbf{51.5} & \textbf{\underline{81.2}} \\
& CoLab & 6 & 46.1 & \textbf{50.2} & 49.7 & 88.9 \\
\hline
\multirow{7}{*}{\tabincell{c}{Brain tumor~\cite{shapey2019artificial}}}& w/o tissue masks & 1 & 84.3 & 91.2 & 80.5 & 15.2 \\
& $K$-means~\cite{arthur2006k} & 6 & 85.0 & 91.3 & 80.3 & 9.4 \\
& Dilated masks~\cite{muller2019does} & 2 & 84.8 & 91.6 & 81.1 & 8.8 \\
& CoLab & 2 & 85.2 & 90.4 & 82.5 & 7.8 \\
& CoLab & 4 & \textbf{\underline{89.0}} & \textbf{91.7} & \textbf{\underline{86.7}} & \textbf{\underline{1.4}} \\
& CoLab & 6 & \textbf{87.9} & \textbf{\underline{92.0}} & \textbf{84.9} & \textbf{2.5} \\
\cline{2-7}
& w/ tissue masks~\cite{shapey2019artificial,ledig2015robust} &  6 & 88.2 & 90.9 & 86.2 & 3.1 \\
\hline
\multirow{6}{*}{\tabincell{c}{Brain lesion~\cite{liew2018large}}}& w/o context labels & 1 & 58.5 & \textbf{68.2} & 58.6 & 51.7 \\
& $K$-means~\cite{arthur2006k} & 6 & 60.0 & 60.7 & \textbf{\underline{70.5}} & \textbf{\underline{19.5}} \\
& Dilated masks~\cite{muller2019does} & 2 & 61.9 & \textbf{\underline{69.8}} & 63.3 & 35.3 \\
& CoLab & 2 & \textbf{\underline{62.8}} & \textbf{68.2} & \textbf{67.4} & 28.1 \\
& CoLab & 4 & \textbf{62.6} & 68.1 & 65.4 & 25.1 \\
& CoLab & 6 & 61.7 & 66.7 & 65.7 & \textbf{24.5} \\
\hline
\end{tabular}
\end{lrbox}
\scalebox{1}{\usebox{\tablekits}}
\end{table*}

\begin{table*}[t]
\centering
\caption{Evaluation of pancreas and pancreatic tumor mass segmentation with different types of context label generation approaches. The best and second best results are in \textbf{bold} with the best also \underline{\textbf{underlined}}.}\label{tabpancreasresults}
\newsavebox{\tablepancreas}
\begin{lrbox}{\tablepancreas}

\begin{tabular}{m{30mm}<{\centering}|m{5mm}<{\centering}|m{10mm}<{\centering}|m{10mm}<{\centering}|m{10mm}<{\centering}|m{10mm}<{\centering}|m{10mm}<{\centering}|m{10mm}<{\centering}|m{10mm}<{\centering}|m{10mm}<{\centering}}
\hlineB{3}
\multirow{2}{*}{Model} & \multirow{2}{*}{$t$} & \multicolumn{4}{c|}{Pancreas} & \multicolumn{4}{c}{Tumor mass} \\ 
& & DSC & SEN & PRC & HD & DSC & SEN & PRC & HD
\\
\hlineB{1}
\hline
w/o context labels & 1 & 82.9 & 82.9 & 85.1 & 13.9 & 48.9 & \textbf{49.2} & 59.3 & \textbf{52.0} \\
$K$-means~\cite{arthur2006k} & 2 & 82.1 & 82.3 & 84.6 & 14.6 & 44.8 & 45.4 & 54.8 & 64.4 \\
Dilated masks~\cite{muller2019does} & 2 & \textbf{\underline{83.5}} & \textbf{\underline{84.6}} & 84.4 & 11.4 & 48.1 & 48.2 & 58.0 & 58.2 \\
CoLab & 2 & \textbf{83.3} & \textbf{83.3} & \textbf{85.3} & \textbf{7.7} & \textbf{\underline{51.1}} & \textbf{\underline{50.8}} & \textbf{\underline{63.2}} & \textbf{\underline{44.8}} \\
CoLab & 4 & 82.7 & 81.7 & \textbf{\underline{86.0}} & \textbf{\underline{7.3}} & 48.7 & 45.9 & \textbf{62.0} & 57.3 \\
CoLab & 6 & 82.7 & 82.9 & 84.7 & 8.0 & \textbf{49.6} & 48.4 & 61.5 & 52.8 \\
\hline
\end{tabular}
\end{lrbox}
\scalebox{1}{\usebox{\tablepancreas}}
\end{table*}

\begin{figure*}[t]
\centering
\includegraphics[width=\textwidth]{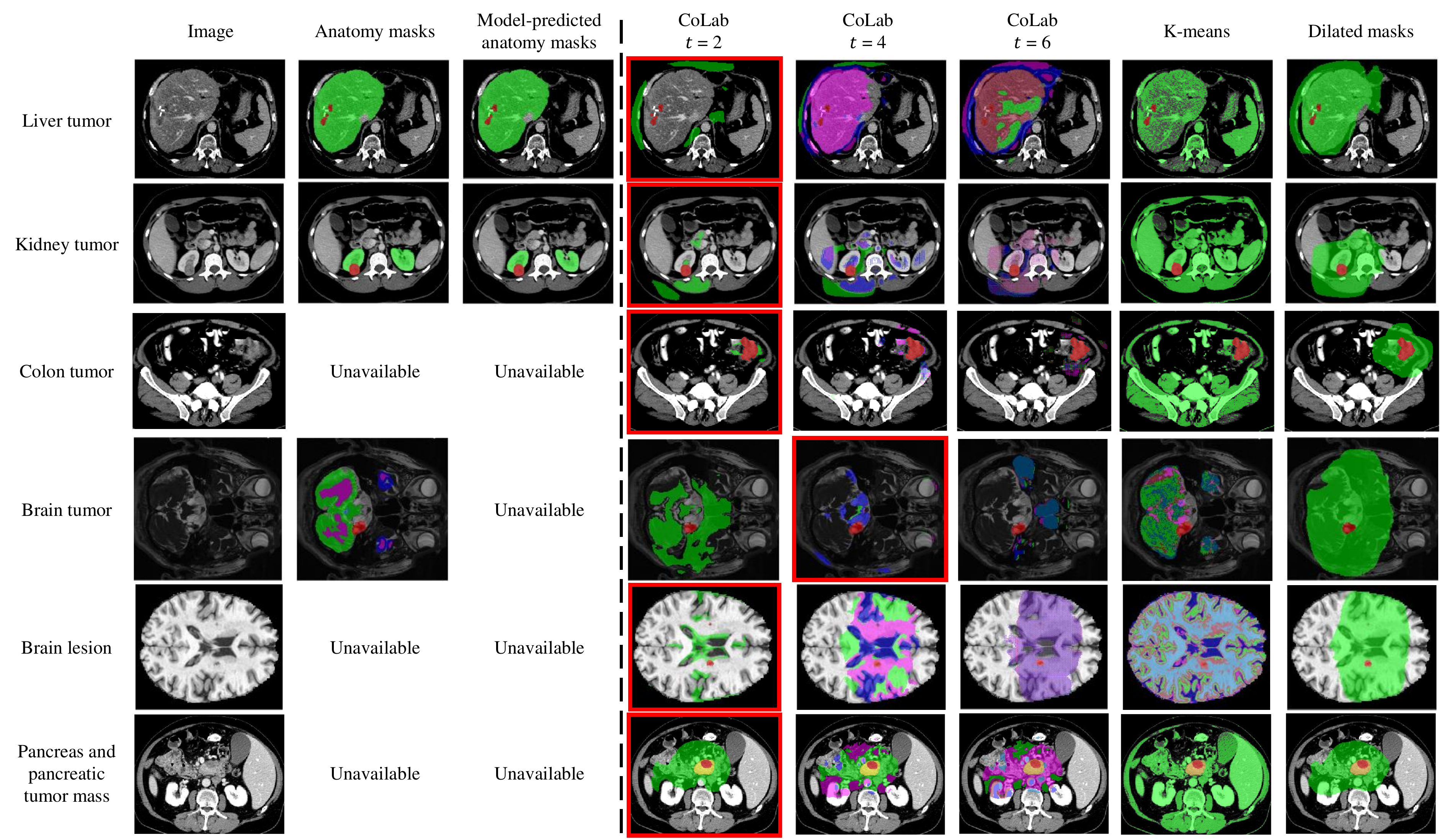}

\caption{Examples of the results $\tilde{f}_{\theta}(\boldsymbol{x}_i)$ of training data with different types of context labels. The binary ROI masks are shown in red, while multi-class ROI masks are shown in red and orange. The context labels are visualized using other colors The context labels which benefit ROI segmentation most and generated without human input are marked with red boxes. CoLab learns to generate specific structure masks for different tasks. This helps the segmentation network to learn better feaure representations for the background class. } \label{fig_contexttask}
\end{figure*}

We summarize the quantitative results of the models trained with different types of context labels on five binary segmentation datasets in Table~\ref{tabresults} and one multi-class segmentation dataset in Table~\ref{tabpancreasresults}. We separate the results according to whether expert knowledge and manual labeling efforts are needed. Taking the manual segmentation results as the ground truth, we calculate DSC, sensitivity (SEN), precision (PRC) and HD. We show some segmentation results with CoLab in Fig.~\ref{fig_segexample}(c) and also summarize all the corresponding context labels in Fig.~\ref{fig_contexttask}. We provide additional examples of context labels in supplementary material.

As discussed in Section~\ref{sec:analysis}, the baseline segmentation model seems to underfit the heterogeneous background samples resulting in many FP and decreased precision. We find most evaluated context labels approaches can improve the models overall segmentation accuracy with increased precision. We observe that adding contextual labels consistently benefits model training and improves performance.

\subsubsection{Performance of baseline methods}

$K$-means based context labels divide the background samples into different classes based on pixel intensity. Those context labels seem to help the segmentation model separating the abdominal organs and bones from background such as fat and air for the abdominal CT images while in the case of brain MR, it may help segmenting the images into different regions such as white matter and ventricles. We find these $k$-means based context labels are effective for liver and kidney tumor segmentation, while being less effective for colon tumor, brain tumor, brain lesion and pancreas tumor segmentation. This may be due to the distinct intensities of large organs and background in CT where the generated labels have meaningful semantic information, benefiting the model training by enhancing the context representations of kidney and liver tumor. Because the appearance of colon tumor is heterogeneous, its segmentation depends more on spatial information and cannot benefit from purely intensity based context labels. As the pancreas label already provides enriched context information for pancreatic tumor segmentation, we find baseline context label techniques such as $k$-means and dilated masks cannot provide additional benefits. The structure of the brain anatomy in MR is more complex and the classes generated by $k$-means would be noisy and disconnected. Here, the $k$-means generated labels are less beneficial for as they provide insufficient context for the ROI.

The context labels based on dilated masks make the segmentation model aware of regions neighbouring the ROIs. We find those context labels are effective for liver tumor, kidney tumor, colon tumor, brain lesion and pancreas segmentation but yield limited improvements for brain tumor segmentation. This may be because liver tumors, kidney tumors, colon tumors and brain lesions appear in varying parts of the primary organ. If trained as binary tasks, the segmentation model predicts FP outside the liver, kidney and colon regions or in anatomically unrelated parts of the brain because these samples share similar characteristics with the ROIs, as shown in Fig.~\ref{fig_segexample}(a). In this case, the dilated masks could enhance the spatial relationship between the ROI and the surrounding context. For example, the dilated masks would connect the liver tumor segmentation with liver context and avoid FP outside the liver regions. However, the samples of VS dataset only contain one ROI per case and VS always appear around the superior vestibular area. Therefore, the segmentation model does not need more spatial information to locate the ROIs and does not benefit from the dilated masks.

As shown in the third column of Fig.~\ref{fig_contexttask}, we obtain high-quality automatic segmentations of liver and kidney regions as the segmentation of large organs is relatively easy and robust in CT images. The model-predicted masks show similar improvements compared to using the manually labeled anatomy masks. However, this approach is not applicable when we do not have prior knowledge available about the ROIs or when we do not have access to external, manually labelled datasets to train supervised models for context label generation.

We summarize the results with component-based post-processing in supplementary material. We find post-processing always decreases the overall performance indicated by DSC for liver tumor, colon tumor and brain lesion segmentation as there always exist multiple separate ROIs in those tasks. For kidney tumor, pancreas tumor and brain tumor segmentation, we find that post-processing is effective in improving the overall segmentation performance of most results while the models trained with context labels are always better than the ones trained with binary tasks. This indicates that the FP next to ROIs cannot be eliminated with simple post-processing.

\subsubsection{CoLab}

As illustrated in Fig.~\ref{fig_segexample}(c), CoLab can effectively reduce different types of false positive predictions and improve precision while preserving sensitivity. CoLab can be applied to different tasks with similar hyper-parameters and show consistent improvements (+2.2 to +8.1 points of DSC). We find CoLab yields better segmentation results than all the other context label approaches that do not use expert knowledge. Moreover, CoLab shows similar improvements compared to using organ masks for liver and kidney tumor segmentation and even better performance for brain tumor segmentation, compared with using tissue masks. It is particularly interesting to find that CoLab can improve the segmentation performance of pancreatic tumor mass, on the top of its strong context label of pancreas. It indicates that there is still room to improve the context representation for segmentation tasks with human defined context labels. CoLab is a flexible and generic method and can be directly applied to boost performance for a variety of segmentation tasks with heterogeneous background classes. We further analyze the context labels generated by CoLab based on intensity histograms and summarize the results in supplementary material. We find the context labels generated by CoLab can highlight the regions which have different intensity distributions from the background samples. The intensity of one of the generated context labels is always similar to that of ROIs. This might indicates that CoLab can generate useful and semantic context labels which can help the segmentation model better distinguish ROIs from similar background samples.

We evaluate CoLab with different number context classes $t$. We find that when the segmentation model tends to make FP far from the ROI such as liver tumor, kidney tumor, colon tumor, brain lesion and pancreas tumor, $t=2$ is the most effective. This may indicate that the segmentation would need the context labels to provide mostly spatial information and $t=2$ makes CoLab focus on the location of the ROI as the task generator only needs to separate one class of context from the remaining background. This single class would enhance the spatial relationship between the ROI and its surrounding objects. This is consistent with the observations of the results as shown in Fig.~\ref{fig_contexttask}. Specifically, the context labels generated with $t=2$ highlight the region next to the liver, kidney and colon regions and could implicitly help the segmentation model focus on the vicinity of the ROI boundaries.

When the baseline segmentation model is making FP around the ROIs such as in the case of VS segmentation, $t=4$ and $t=6$ seem more effective. This is probably because the segmentation requires more structured information of surrounding objects and the complicated context of brain anatomy could be better represented when divided into more than 2 classes. As illustrated in Fig.~\ref{fig_contexttask}, context labels generated with $t=4$ highlight structure information around the ROIs. Specifically, CoLab generates segmentations of the superior vestibular area for VS segmentation where the baseline had predicted FP. Similarly, with the task of brain lesion segmentation, the context labels highlight the regions around ventricles to avoid FP in this area. By doing so, the model learns spatial information by building specific representations for the ventricle regions. 

We should note that CoLab may not be very effective with large $t$, specifically we find CoLab with $t=6$ would not bring much improvements for kidney tumor, colon tumor and brain lesion segmentation. As shown in Fig.~\ref{fig_contexttask}, we observe that the generated context labels would yield checkerboard patterns. This might be because further decomposing the background class cannot make the task easier to learn. In this case, we find the task generator cannot generate coherent labels and produce similar probabilities for the different classes in those regions, making the context labels trivial and hard to learn.

In general, we find $t=2$ is the most effective choice for most cases in terms of DSC, especially for tumor segmentation in abdominal CT images. This might indicate that the background representation is not so complex and could be improved by adding refined spatial information. When trained with $t$ larger than 2, the models could have better results in terms of HD, compared with the models trained with $t=2$. In this case, the models learn the background class better, as they make less FP away from the ROIs. However, choosing $t$ larger than 2 seems to make the models fail to learn the foreground class well, resulting in worse DSC and sensitivity. This may be because increasing $t$ could make the prediction of foreground class more difficult.

We suggest the use of $t=2$ when the baseline model makes FP distant from the ROIs, and $t=4$ when over-segmentation is observed closer to the ROIs. In addition, we look into the intensity histograms of ROIs for different datasets and summarize the results in supplementary material. We find the intensity distributions of most ROIs are not distinct from that of the background samples, apart from brain tumor. Perhaps for brain tumor segmentation, the model can benefit from knowing the regions which share the most similar intensity distributions with the ROIs, thus requiring more classes of context labels. This can be a potential factor due to which the optimal class of context labels $t$ for brain tumor is 4 instead of 2.

\subsection{Effect on logit distributions}

We also show the effect of CoLab ($t=2$ for liver and kidney tumor segmentation and $t=4$ for brain tumor segmentation) on the logit distributions in Fig.~\ref{fig_logitmap}(c, f, i). We find CoLab has a similar regularization effect than using anatomy masks. By decomposing the background class, the segmentation model can robustly map the background samples to different locations away from the decision boundary, reducing the logit shifts, and thus yielding fewer FP.

\section{Conclusion}
\label{sec:conclusion}

In this study, we find that segmentation models are prone to over-segment the ROIs in the presence of heterogeneous background classes. With the observation of network behaviour across multiple segmentation tasks, we conclude that this is due to the model being unable to learn discriminative representations from the background samples and the models can be significantly improved by incorporating context labels during training. We present CoLab, a generic method to automatically generate effective context labels through a meta-learning scheme. We show CoLab improves overall segmentation performance substantially for several challenging segmentation tasks, without the need of expert knowledge and laborious labeling efforts. In future work, we will evaluate CoLab in settings such as domain shifts. It will also be interesting to explore automatic context label generation in multi-task settings where a shared backbone model is trained to automatically identify interesting subclasses across tasks and even datasets.



\section*{Acknowledgements}
Z.Li is grateful for the China Scholarship Council (CSC) Imperial Scholarship. This project has received funding from the ERC under the EU's Horizon 2020 research and innovation programme (grant No. 757173). C.Chen and C.Ouyang were supported by the EPSRC Programme Grant
(EP/P001009/1) and the UKRI Innovate UK Grant (No.104691).

\section*{Supplementary Material}

\subsection{Network behaviour with the training data}

\begin{figure*}[t]
\centering
\includegraphics[width=\textwidth]{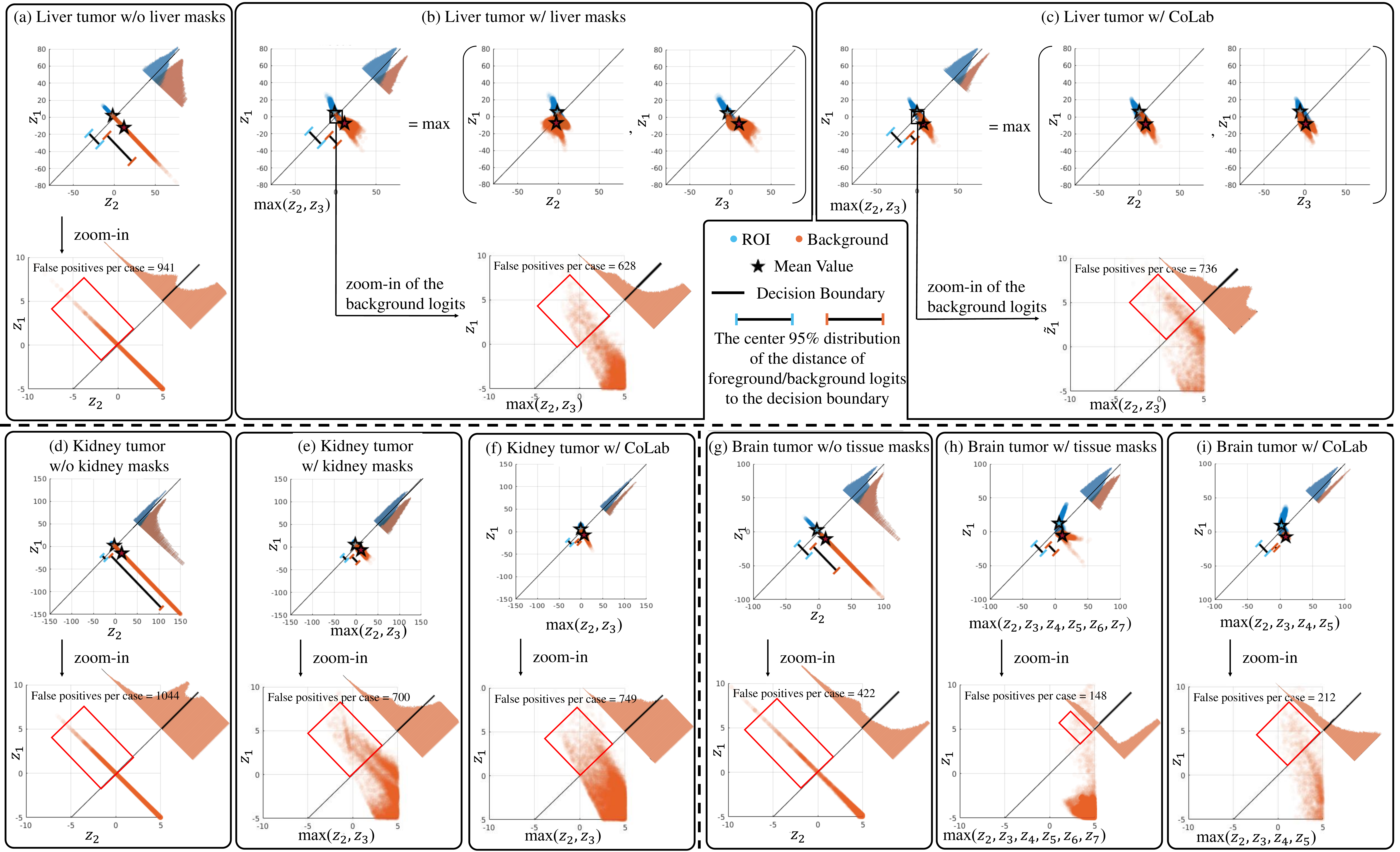}

\caption{Activations of the classification layer (logit $z_1$ for ROI, other logits for background) when processing the ROI (blue) or background sample (orange) of training data. It seems difficult for a CNN to learn good representations over heterogeneous background samples due to the varying characteristics leading to diverse feature embeddings. Extending the label space with anatomy masks or CoLab can help CNN reduce false positives and fit the training data better.} \label{fig_logitmaptraining}
\end{figure*}

We show the network behaviour with the training data in Fig.~\ref{fig_logitmaptraining}. The behaviour is similar to the ones with test data. We observe little performance gap between the training and test data. This indicates that the reason why the model cannot recognize the background sample well is that it underfits the training data. In consequence, the model is likely to make false positives even for the training data. We find anatomy masks and CoLab both can alleviate this issue.

\subsection{Implementation details}

\subsubsection{Derivatives calculation based on implicit function theorem}

One can obtain $\frac{\partial \theta^*}{\partial \omega}$ based on implicit function theorem~\cite{bengio2000gradient}. In order to compute $\frac{\partial \theta^*}{\partial \omega}$, one can assume that $\nabla_\theta  \mathcal{L}_{seg}(\theta^*,\omega)$ is continuously differentiable around $\boldsymbol{0}$ w.r.t. $\omega$ following~\cite{zela2019understanding}, and calculate the total derivative of $\nabla_\theta  \mathcal{L}_{seg}(\theta^*,\omega)=\boldsymbol{0}$ w.r.t. $\omega$ from both sides:

\begin{equation}
    \frac{\partial \nabla_{\theta}\mathcal{L}_{seg}(\theta^*,\omega)}{\partial \theta}\frac{\partial \theta^*}{\partial \omega} + \frac{\partial \nabla_{\theta}\mathcal{L}_{seg}(\theta^*,\omega)}{ \partial \omega} = \boldsymbol{0}.
    \label{eq:funcimplicit}
\end{equation}

With the assumption that $\nabla^2_{\theta}\mathcal{L}_{seg}(\theta^*,\omega)$ is invertible, one can yield:

\begin{equation}
    \frac{\partial \theta^*}{\partial \omega} = -(\nabla^2_{\theta}\mathcal{L}_{seg}(\theta^*,\omega))^{-1} \nabla^2_{\theta,\omega}\mathcal{L}_{seg}(\theta^*,\omega),
    \label{eq:omegacal}
\end{equation}

However, the Hassian $\nabla^2_{\theta}\mathcal{L}_{seg}(\theta^*,\omega)$ is hard to compute. Therefore, in this study we follow some heuristics such as~\cite{finn2017model} to approximate $\frac{\partial \theta^*}{\partial \omega}$.

\subsubsection{Meta-gradient calculation}

In our study, we further simplify the calculation of Eq.~\ref{eq:func3s} with the finite difference approximation following~\cite{liu2018darts}. We calculate network parameters $\theta^{\pm} = \theta \pm \epsilon\nabla_{\theta}\mathcal{L}_{ROI}(\theta^*)$ given some small scalars $\epsilon = 0.01 / \left\Vert \nabla_{\theta}\mathcal{L}_{ROI}(\theta^*) \right\Vert_2$.
With this notion, the second-order gradient can be written as:

\begin{equation}
\begin{aligned}
    & \nabla_{\omega^t,\theta}^{2} \mathcal{L}_{seg}(\theta^*,\omega^t) \approx \frac{\nabla_{\omega^t} \mathcal{L}_{seg}(\theta^{+}, \omega^t) - \nabla_{\omega^t} \mathcal{L}_{seg}(\theta^{-}, \omega^t)}{2\epsilon\nabla_{\theta}\mathcal{L}_{ROI}(\theta^*)} \\
    & \Leftrightarrow \quad \quad \nabla_{\omega^t,\theta}^{2} \mathcal{L}_{seg}(\theta^*,\omega^t) \nabla_{\theta}\mathcal{L}_{ROI}(\theta^*) \approx \\
    & \frac{\nabla_{\omega^t} \mathcal{L}_{seg}(\theta^{+}, \omega^t) - \nabla_{\omega^t} \mathcal{L}_{seg}(\theta^{-}, \omega^t)}{2\epsilon}.
    \label{eqn:dartsfinitediffs}
\end{aligned}
\end{equation}

In this way, we reduce the calculation complexity from $O(|\omega||\theta|)$ to $O(|\omega|+|\theta|)$ and can approximate Eq.~\ref{eq:func3s} with two forward processes of $\tilde{f}_{\theta^*}$.

\subsubsection{Update of task generator}

We find the segmentation model would not learn well if the context labels changes too frequently. Therefore, we only update the task generator once several epochs. Specifically, we only update $\omega$ following step 8 in Algorithm~\ref{alg:CoLab} once 10 epochs. Another benefit of updating $\omega$ once several epochs is that we do not need much more time than the vanilla training process. In practice, we find CoLab would just add extra 20\% to 30\% of the vanilla training time.

\subsection{Hyper-parameters of CoLab}

We summarize the hyper-parameters we use in Table~\ref{tabhyperp}. We choose similar hyper-parameters of the soft dilated masks $\mathcal{M}_i$ for all tasks. We find that the generated context labels for kidney tumor segmentation would change a lot across different training epochs, therefore we choose to update $g_\omega$ for this task less frequently. Similarly, we find that it is relatively hard to optimize CoLab for colon tumor segmentation, as the generated context labels would easily degrade to one class. Thus, we also update $g_\omega$ for this task less frequently. In general, we find the same set of hyper-parameters can be effective across varied datasets, indicating that CoLab is robust and flexible for different application scenarios. 

\begin{table}[t]
\centering
\caption{Hyper-parameters of CoLab for different datasets.}\label{tabhyperp}
\newsavebox{\tablehyperp}
\begin{lrbox}{\tablehyperp}

\begin{tabular}{m{30mm}<{\centering}|m{10mm}<{\centering}|m{10mm}<{\centering}|m{20mm}<{\centering}}
\hlineB{3}

Task & $m$ & $\tau$ & Update period \\
\hline

Liver tumor & 30 & 20 & 5 iterations \\
Kidney tumor & 30 & 20 & 20 iterations \\
Colon tumor & 10 & 20 & 100 iterations \\
Brain tumor & 30 & 20 & 10 iterations \\
Brain lesion & 30 & 20 & 10 iterations \\
Pancreas and pancreatic tumor mass$^\dagger$ & 10$^\ddagger$ & 20 & 10 iterations
\\
\hlineB{1}

\end{tabular}
\end{lrbox}
\scalebox{1}{\usebox{\tablehyperp}}

{\raggedright \quad $^\dagger$We calculate $\mathcal{L}_c$ using BCE only on the pancreas class. \par}

{\raggedright \quad $^\ddagger$We generate the soft dilated masks by taking the pancreas and pancreatic tumor mass as a whole. \par}

\end{table}

\subsection{Results with post-processing}

We summarize all the results with component-based post-processing~\cite{isensee2021nnu} in Table~\ref{tabportp}. We find this post-processing would always decrease the segmentation performance for liver tumor, colon tumor and brain lesion as there always exist multiple ROIs per case for those tasks. For kidney tumor, brain tumor and pancreas tumor segmentation, post-processing can improve overall segmentation performance with higher precision for most cases. After post-processing, we find CoLab can help the model achieve the best segmentation performance when compared with other context label generation approaches without human efforts. 

\begin{table*}[t]
\centering
\caption{Evaluation of different segmentation tasks with different types of context labels approaches. The results are calculated with component-based post-processing. We divide the results according to whether expert knowledge and labeling efforts are included. The best and second best results without human efforts are in \textbf{bold} with the best also \underline{\textbf{underlined}}.}\label{tabportp}
\newsavebox{\tablepostp}
\begin{lrbox}{\tablepostp}

\begin{tabular}{m{30mm}<{\centering}|m{50mm}<{\centering}|m{5mm}<{\centering}|m{10mm}<{\centering}|m{10mm}<{\centering}|m{10mm}<{\centering}|m{10mm}<{\centering}}
\hlineB{3}
Task & Method & $t$ & DSC & SEN & PRC & HD
\\
\hlineB{1}
\multirow{8}{*}{\tabincell{c}{Liver tumor~\cite{bilic2019liver}}}& w/o liver masks & 1 & 37.3 & 29.5 & 61.6 & 121.0 \\
& K-means~\cite{arthur2006k} & 2 & \textbf{45.7} & \textbf{37.5} & 69.9 & 111.8 \\
& Dilated masks~\cite{muller2019does} & 2 & \textbf{45.7} & 37.3 & \textbf{\underline{71.5}} & 108.8 \\
& CoLab & 2 & \textbf{\underline{48.0}} & \textbf{\underline{40.7}} & \textbf{70.3} & 116.8 \\
& CoLab & 4 & 41.6 & 35.2 & 63.7 & \textbf{101.8} \\
& CoLab & 6 & 44.0 & 37.2 & 66.7 & \textbf{\underline{97.5}} \\
\cline{2-7}
& w/ model-predicted liver masks~\cite{xu2015efficient} & 2 & 46.2 & 37.6 & 73.4 & 99.0 \\
& w/ liver masks~\cite{bilic2019liver} & 2 & 44.8 & 36.0 & 70.2 & 111.4 \\
\hline
\multirow{8}{*}{\tabincell{c}{Kidney tumor~\cite{heller2019kits19}}}& w/o kidney masks & 1 & 78.2 & \textbf{79.2} & \textbf{79.8} & 30.0 \\
& K-means~\cite{arthur2006k} & 2 & \textbf{78.3} & \textbf{79.2} & 79.3 & 24.2 \\
& Dilated masks~\cite{muller2019does} & 2 & 77.5 & 78.6 & 77.7 & 24.2 \\
& CoLab & 2 & \textbf{\underline{80.5}} & \textbf{\underline{81.0}} & \textbf{\underline{81.0}} & \textbf{\underline{17.6}} \\
& CoLab & 4 & 76.1 & 76.5 & 78.6 & \textbf{22.7} \\
& CoLab & 6 & 75.5 & 76.9 & 76.2 & 37.1 \\
\cline{2-7}
& w/ model-predicted kidney masks~\cite{xu2015efficient} & 2 & 79.5 & 78.9 & 84.7 & 16.0 \\
& w/ kidney masks~\cite{heller2019kits19} & 2 & 80.9 & 81.5 & 82.3 & 17.6 \\
\hline
\multirow{6}{*}{\tabincell{c}{Colon tumor~\cite{simpson2019large}}}& w/o context labels & 1 & \textbf{\underline{48.0}} & \textbf{\underline{48.8}} & 52.5 & \textbf{52.6} \\
& K-means~\cite{arthur2006k} & 2 & 44.3 & 41.9 & 50.7 & 75.2 \\
& Dilated masks~\cite{muller2019does} & 2 & 46.6 & 46.4 & \textbf{54.1} & 61.5 \\
& CoLab & 2 & \textbf{47.3} & \textbf{46.9} & 52.4 & 53.3 \\
& CoLab & 4 & 44.6 & 43.5 & \textbf{\underline{54.4}} & \textbf{\underline{51.3}} \\
& CoLab & 6 & 45.1 & 45.6 & 49.8 & 58.0 \\
\hline
\multirow{7}{*}{\tabincell{c}{Brain tumor~\cite{shapey2019artificial}}}& w/o tissue masks & 1 & 84.3 & 87.4 & 82.0 & 5.1 \\
& K-means~\cite{arthur2006k} & 6 & 86.9 & 91.3 & 83.4 & 1.6 \\
& Dilated masks~\cite{muller2019does} & 2 & 83.1 & 85.7 & 80.9 & 4.6 \\
& CoLab & 2 & 84.1 & 86.5 & 82.4 & 4.2 \\
& CoLab & 4 & \textbf{\underline{89.0}} & \textbf{91.7} & \textbf{\underline{86.7}} & \textbf{\underline{1.4}} \\
& CoLab & 6 & \textbf{88.6} & \textbf{\underline{92.0}} & \textbf{86.0} & \textbf{1.5} \\
\cline{2-7}
& w/ tissue masks~\cite{shapey2019artificial,ledig2015robust} &  6 & 88.7 & 90.9 & 87.0 & 1.4 \\
\hline
\multirow{6}{*}{\tabincell{c}{Brain lesion~\cite{liew2018large}}}& w/o tissue masks & 1 & 57.8 & 57.1 & 68.1 & 25.4 \\
& K-means~\cite{arthur2006k} & 6 & 58.6 & 56.2 & \textbf{\underline{74.6}} & 24.6 \\
& Dilated masks~\cite{muller2019does} & 2 & \textbf{\underline{61.8}} & \textbf{\underline{62.2}} & 67.6 & \textbf{\underline{21.6}} \\
& CoLab & 2 & 61.1 & 60.7 & \textbf{72.0} & 23.6 \\
& CoLab & 4 & \textbf{61.4} & \textbf{61.8} & 70.0 & \textbf{23.0} \\
& CoLab & 6 & 59.4 & 58.9 & 69.4 & \textbf{\underline{21.6}} \\
\hline
\end{tabular}
\end{lrbox}
\scalebox{1}{\usebox{\tablepostp}}
\end{table*}

\begin{table*}[t]
\centering
\caption{Evaluation of pancreas and pancreatic tumor mass segmentation with different types of context label generation approaches. The results are calculated with component-based post-processing. The best and second best results are in \textbf{bold} with the best also \underline{\textbf{underlined}}.}\label{tabpancreasresultspostp}
\newsavebox{\tablepancreaspostp}
\begin{lrbox}{\tablepancreaspostp}

\begin{tabular}{m{30mm}<{\centering}|m{5mm}<{\centering}|m{10mm}<{\centering}|m{10mm}<{\centering}|m{10mm}<{\centering}|m{10mm}<{\centering}|m{10mm}<{\centering}|m{10mm}<{\centering}|m{10mm}<{\centering}|m{10mm}<{\centering}}
\hlineB{3}
\multirow{2}{*}{Model} & \multirow{2}{*}{$t$} & \multicolumn{4}{c|}{Pancreas} & \multicolumn{4}{c}{Tumor mass} \\ 
& & DSC & SEN & PRC & HD & DSC & SEN & PRC & HD
\\
\hlineB{1}
\hline
w/o context labels & 1 & 82.3 & 81.9 & 85.0 & 9.9 & 49.0 & \textbf{48.5} & 59.5 & \textbf{47.6} \\
K-means~\cite{arthur2006k} & 2 & 83.0 & 82.3 & \textbf{\underline{86.1}} & \textbf{7.5} & 47.2 & 45.4 & 58.3 & 52.6 \\
Dilated masks~\cite{muller2019does} & 2 & \textbf{\underline{83.7}} & \textbf{\underline{84.1}} & 85.4 & 8.1 & 49.7 & 48.0 & 60.1 & 55.1 \\
CoLab & 2 & \textbf{83.3} & \textbf{83.3} & 85.5 & \textbf{\underline{7.1}} & \textbf{\underline{51.6}} & \textbf{\underline{50.8}} & \textbf{\underline{63.9}} & \textbf{\underline{41.9}} \\
CoLab & 4 & 82.3 & 81.3 & \textbf{85.8} & 8.1 & 47.9 & 45.2 & 60.7 & 61.5 \\
CoLab & 6 & 82.7 & 82.9 & 84.8 & 8.1 & \textbf{49.8} & 48.4 & \textbf{61.7} & 53.7 \\
\hline
\end{tabular}
\end{lrbox}
\scalebox{1}{\usebox{\tablepancreaspostp}}
\end{table*}

\subsection{Additional visualization of context labels}

\begin{figure*}[t]
\centering
\includegraphics[width=\textwidth]{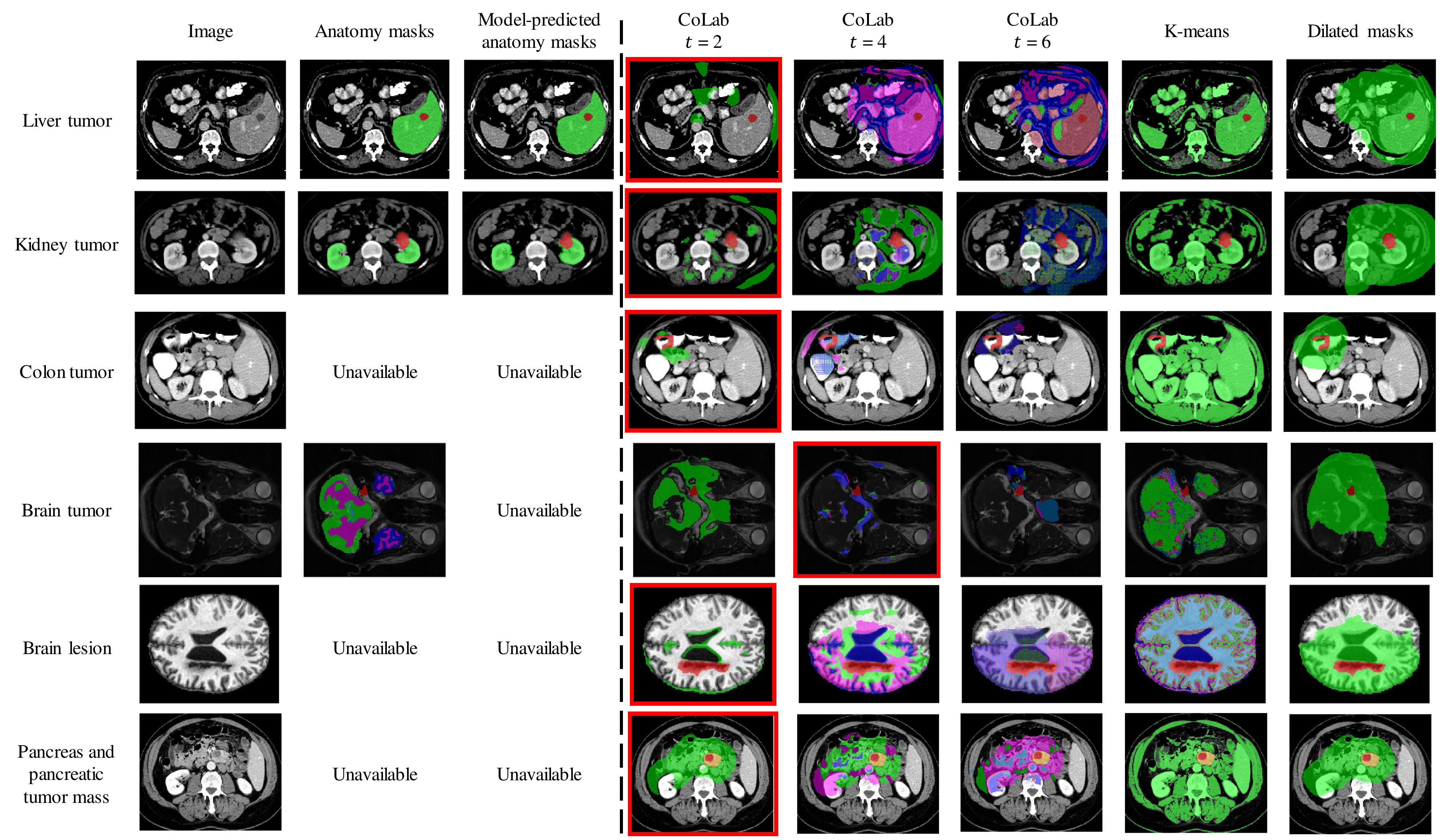}

\caption{Additional examples of the results $\tilde{f}_{\theta}(\boldsymbol{x}_i)$ of training data with different types of context labels. The binary ROI masks are shown in red, while multi-class ROI masks are shown in red and orange. The context labels are visualized using other colors. The context labels which benefit ROI segmentation most and generated without human input are marked with red boxes.} \label{fig_contexttask2}
\end{figure*}

We provide additional examples of generated tasks in Fig.~\ref{fig_contexttask2}, which is a supplement to Fig.~\ref{fig_contexttask}.

\subsection{Intensity histograms of ROIs and context labels}

\begin{figure*}[t]
\centering
\includegraphics[width=\textwidth]{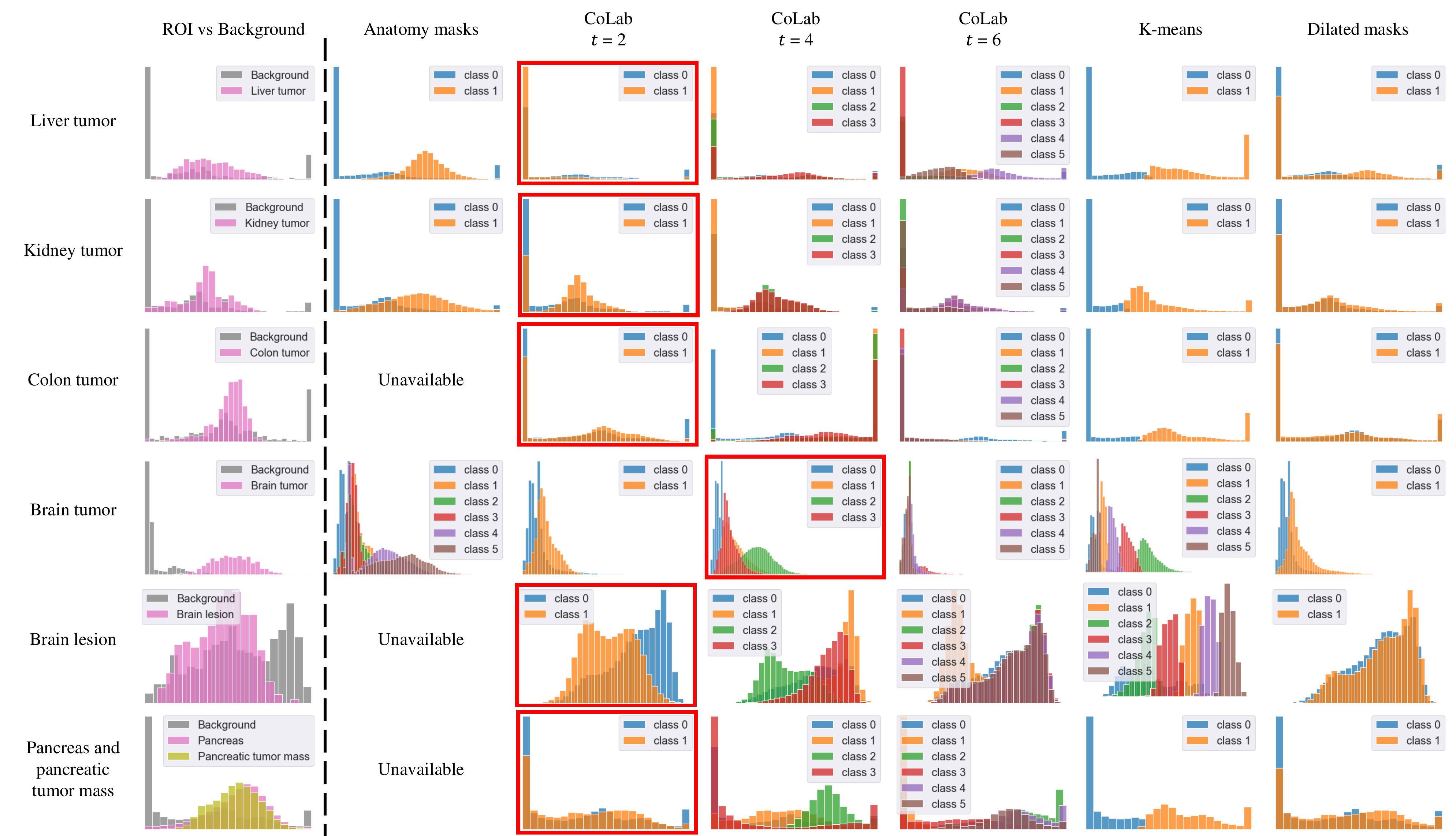}

\caption{The intensity histogram of ROIs calculated based on $\boldsymbol{y}_i$ and the different context labels calculated based on $\tilde{f}_{\theta}(\boldsymbol{x}_i)$. The histograms of context labels which benefit ROI segmentation most and generated without human input are marked with red boxes.} \label{fig_hist}
\end{figure*}

We analyze the intensity of context labels of different datasets and summarize the intensity histograms in right part of Fig.~\ref{fig_hist}. K-means based context labels are generated based on pixel intensity, therefore the histograms of different classes generally distribute in separate regions. Different classes of dilated masks based context labels always share similar intensity distributions, as dilated masks only takes spatial information into account and does not contain much semantic meanings. CoLab learns to generate context labels within the dilated regions based on meta-learning. Similar to anatomy masks, CoLab would highlight related semantic regions which similar to ROIs but differ from other background samples. We should note that here we only analyze the context labels learned in the final models for CoLab. As the task generator would also be optimized during the training process, it is hard to interpret how CoLab help the segmentation model learn better.

We also look into the intensity histograms of ROIs in left part of Fig.~\ref{fig_hist}. We find the intensity histograms of ROIs always largely overlap with that of background samples. The model might require external spatial information to enhance the prior knowledge of ROI location. $t=2$ could make CoLab highlight the surrounding regions which can help the model reduce false positive predictions on samples with similar intensity but distant from ROIs. In this case, the model cannot benefit from more detailed class representation with $t>2$ which would further focus on decomposing the samples around ROIs. On the contrary, we notice only the intensity distributions of brain tumor would differentiate from that of background samples. In order to further improve the segmentation accuracy, the model might require more specific subregion information to discriminate between ROIs and the regions with similar intensity. This can be a potential factor due to which only the optimal context class $t$ of brain tumor segmentation is 4 instead of 2. Practitioners could utilize the intensity histograms of ROI to choose the optimal $t$. According to these findings, we suggest to choose $t = 2$ for ROIs having similar intensity with the background samples, while choose $t > 2$ for ROIs with distinct intensity distributions from the background samples.

\bibliographystyle{ieee}
\bibliography{reference}

\begin{thebibliography}{10}\itemsep=-1pt

\bibitem{arthur2006k}
D.~Arthur and S.~Vassilvitskii.
\newblock k-means++: The advantages of careful seeding.
\newblock Technical report, Stanford, 2006.

\bibitem{attiyeh2018survival}
M.~A. Attiyeh, J.~Chakraborty, A.~Doussot, L.~Langdon-Embry, S.~Mainarich,
  M.~G{\"o}nen, V.~P. Balachandran, M.~I. D’Angelica, R.~P. DeMatteo, W.~R.
  Jarnagin, et~al.
\newblock Survival prediction in pancreatic ductal adenocarcinoma by
  quantitative computed tomography image analysis.
\newblock {\em Annals of surgical oncology}, 25(4):1034--1042, 2018.

\bibitem{attiyeh2019preoperative}
M.~A. Attiyeh, J.~Chakraborty, L.~Gazit, L.~Langdon-Embry, M.~Gonen, V.~P.
  Balachandran, M.~I. D'Angelica, R.~P. DeMatteo, W.~R. Jarnagin, T.~P.
  Kingham, et~al.
\newblock Preoperative risk prediction for intraductal papillary mucinous
  neoplasms by quantitative ct image analysis.
\newblock {\em Hpb}, 21(2):212--218, 2019.

\bibitem{bengio2000gradient}
Y.~Bengio.
\newblock Gradient-based optimization of hyperparameters.
\newblock {\em Neural computation}, 12(8):1889--1900, 2000.

\bibitem{bilic2019liver}
P.~Bilic, P.~F. Christ, E.~Vorontsov, G.~Chlebus, H.~Chen, Q.~Dou, C.-W. Fu,
  X.~Han, P.-A. Heng, J.~Hesser, et~al.
\newblock The liver tumor segmentation benchmark (lits).
\newblock {\em arXiv preprint arXiv:1901.04056}, 2019.

\bibitem{bragman2019stochastic}
F.~J. Bragman, R.~Tanno, S.~Ourselin, D.~C. Alexander, and J.~Cardoso.
\newblock Stochastic filter groups for multi-task cnns: Learning specialist and
  generalist convolution kernels.
\newblock In {\em Proceedings of the IEEE/CVF International Conference on
  Computer Vision}, pages 1385--1394, 2019.

\bibitem{cai2021ace}
J.~Cai, Y.~Wang, and J.-N. Hwang.
\newblock Ace: Ally complementary experts for solving long-tailed recognition
  in one-shot.
\newblock In {\em Proceedings of the IEEE/CVF International Conference on
  Computer Vision}, pages 112--121, 2021.

\bibitem{chakraborty2018ct}
J.~Chakraborty, A.~Midya, L.~Gazit, M.~Attiyeh, L.~Langdon-Embry, P.~J. Allen,
  R.~K. Do, and A.~L. Simpson.
\newblock Ct radiomics to predict high-risk intraductal papillary mucinous
  neoplasms of the pancreas.
\newblock {\em Medical physics}, 45(11):5019--5029, 2018.

\bibitem{finn2017model}
C.~Finn, P.~Abbeel, and S.~Levine.
\newblock Model-agnostic meta-learning for fast adaptation of deep networks.
\newblock In {\em International Conference on Machine Learning}, pages
  1126--1135. PMLR, 2017.

\bibitem{franceschi2018bilevel}
L.~Franceschi, P.~Frasconi, S.~Salzo, R.~Grazzi, and M.~Pontil.
\newblock Bilevel programming for hyperparameter optimization and
  meta-learning.
\newblock In {\em International Conference on Machine Learning}, pages
  1568--1577. PMLR, 2018.

\bibitem{glocker2012joint}
B.~Glocker, O.~Pauly, E.~Konukoglu, and A.~Criminisi.
\newblock Joint classification-regression forests for spatially structured
  multi-object segmentation.
\newblock In {\em European conference on computer vision}, pages 870--881.
  Springer, 2012.

\bibitem{heller2019kits19}
N.~Heller, N.~Sathianathen, A.~Kalapara, E.~Walczak, K.~Moore, H.~Kaluzniak,
  J.~Rosenberg, P.~Blake, Z.~Rengel, M.~Oestreich, et~al.
\newblock The kits19 challenge data: 300 kidney tumor cases with clinical
  context, ct semantic segmentations, and surgical outcomes.
\newblock {\em arXiv preprint arXiv:1904.00445}, 2019.

\bibitem{isensee2021nnu}
F.~Isensee, P.~F. Jaeger, S.~A. Kohl, J.~Petersen, and K.~H. Maier-Hein.
\newblock nnu-net: a self-configuring method for deep learning-based biomedical
  image segmentation.
\newblock {\em Nature methods}, 18(2):203--211, 2021.

\bibitem{islam2021spatially}
M.~Islam and B.~Glocker.
\newblock Spatially varying label smoothing: Capturing uncertainty from expert
  annotations.
\newblock In {\em International Conference on Information Processing in Medical
  Imaging}, pages 677--688. Springer, 2021.

\bibitem{kang2019decoupling}
B.~Kang, S.~Xie, M.~Rohrbach, Z.~Yan, A.~Gordo, J.~Feng, and Y.~Kalantidis.
\newblock Decoupling representation and classifier for long-tailed recognition.
\newblock {\em arXiv preprint arXiv:1910.09217}, 2019.

\bibitem{xu2015efficient}
B.~A. Landman, Z.~Xu, J.~E. Igelsias, M.~Styner, T.~R. Langerak, and A.~Klein.
\newblock 2015 miccai multi-atlas labeling beyond the cranial vault –
  workshop and challenge.
\newblock Accessed Dec. 2020. [Online]. Available:
  \url{https://www.synapse.org/#!Synapse:syn3193805}, doi: 10.7303/syn3193805.

\bibitem{ledig2015robust}
C.~Ledig, R.~A. Heckemann, A.~Hammers, J.~C. Lopez, V.~F. Newcombe,
  A.~Makropoulos, J.~L{\"o}tj{\"o}nen, D.~K. Menon, and D.~Rueckert.
\newblock Robust whole-brain segmentation: application to traumatic brain
  injury.
\newblock {\em Medical image analysis}, 21(1):40--58, 2015.

\bibitem{li2019overfitting}
Z.~Li, K.~Kamnitsas, and B.~Glocker.
\newblock Overfitting of neural nets under class imbalance: Analysis and
  improvements for segmentation.
\newblock In {\em International Conference on Medical Image Computing and
  Computer-Assisted Intervention}, pages 402--410. Springer, 2019.

\bibitem{li2020analyzing}
Z.~Li, K.~Kamnitsas, and B.~Glocker.
\newblock Analyzing overfitting under class imbalance in neural networks for
  image segmentation.
\newblock {\em IEEE Transactions on Medical Imaging}, 40(3):1065--1077, 2020.

\bibitem{li2019deepvolume}
Z.~Li, J.~Yu, Y.~Wang, H.~Zhou, H.~Yang, and Z.~Qiao.
\newblock Deepvolume: Brain structure and spatial connection-aware network for
  brain mri super-resolution.
\newblock {\em IEEE transactions on cybernetics}, 2019.

\bibitem{liew2018large}
S.-L. Liew, J.~M. Anglin, N.~W. Banks, M.~Sondag, K.~L. Ito, H.~Kim, J.~Chan,
  J.~Ito, C.~Jung, N.~Khoshab, et~al.
\newblock A large, open source dataset of stroke anatomical brain images and
  manual lesion segmentations.
\newblock {\em Scientific data}, 5:180011, 2018.

\bibitem{liu2018darts}
H.~Liu, K.~Simonyan, and Y.~Yang.
\newblock Darts: Differentiable architecture search.
\newblock {\em arXiv preprint arXiv:1806.09055}, 2018.

\bibitem{liu2019self}
S.~Liu, A.~J. Davison, and E.~Johns.
\newblock Self-supervised generalisation with meta auxiliary learning.
\newblock {\em arXiv preprint arXiv:1901.08933}, 2019.

\bibitem{liu2019large}
Z.~Liu, Z.~Miao, X.~Zhan, J.~Wang, B.~Gong, and S.~X. Yu.
\newblock Large-scale long-tailed recognition in an open world.
\newblock In {\em Proceedings of the IEEE/CVF Conference on Computer Vision and
  Pattern Recognition}, pages 2537--2546, 2019.

\bibitem{milletari2016v}
F.~Milletari, N.~Navab, and S.-A. Ahmadi.
\newblock V-net: Fully convolutional neural networks for volumetric medical
  image segmentation.
\newblock In {\em 2016 fourth international conference on 3D vision (3DV)},
  pages 565--571. IEEE, 2016.

\bibitem{muller2019does}
R.~M{\"u}ller, S.~Kornblith, and G.~Hinton.
\newblock When does label smoothing help?
\newblock {\em arXiv preprint arXiv:1906.02629}, 2019.

\bibitem{navon2020auxiliary}
A.~Navon, I.~Achituve, H.~Maron, G.~Chechik, and E.~Fetaya.
\newblock Auxiliary learning by implicit differentiation.
\newblock {\em arXiv preprint arXiv:2007.02693}, 2020.

\bibitem{novak2018sensitivity}
R.~Novak, Y.~Bahri, D.~A. Abolafia, J.~Pennington, and J.~Sohl-Dickstein.
\newblock Sensitivity and generalization in neural networks: an empirical
  study.
\newblock {\em arXiv preprint arXiv:1802.08760}, 2018.

\bibitem{paszke2019pytorch}
A.~Paszke, S.~Gross, F.~Massa, A.~Lerer, J.~Bradbury, G.~Chanan, T.~Killeen,
  Z.~Lin, N.~Gimelshein, L.~Antiga, et~al.
\newblock Pytorch: An imperative style, high-performance deep learning library.
\newblock {\em Advances in neural information processing systems},
  32:8026--8037, 2019.

\bibitem{pedregosa2016hyperparameter}
F.~Pedregosa.
\newblock Hyperparameter optimization with approximate gradient.
\newblock In {\em International conference on machine learning}, pages
  737--746. PMLR, 2016.

\bibitem{qaiser2021multiple}
T.~Qaiser, S.~Winzeck, T.~Barfoot, T.~Barwick, S.~J. Doran, M.~F. Kaiser,
  L.~Wedlake, N.~Tunariu, D.-M. Koh, C.~Messiou, et~al.
\newblock Multiple instance learning with auxiliary task weighting for multiple
  myeloma classification.
\newblock In {\em International Conference on Medical Image Computing and
  Computer-Assisted Intervention}, pages 786--796. Springer, 2021.

\bibitem{setio2016pulmonary}
A.~A.~A. Setio, F.~Ciompi, G.~Litjens, P.~Gerke, C.~Jacobs, S.~J. Van~Riel,
  M.~M.~W. Wille, M.~Naqibullah, C.~I. S{\'a}nchez, and B.~Van~Ginneken.
\newblock Pulmonary nodule detection in ct images: false positive reduction
  using multi-view convolutional networks.
\newblock {\em IEEE transactions on medical imaging}, 35(5):1160--1169, 2016.

\bibitem{shapey2019artificial}
J.~Shapey, G.~Wang, R.~Dorent, A.~Dimitriadis, W.~Li, I.~Paddick, N.~Kitchen,
  S.~Bisdas, S.~R. Saeed, S.~Ourselin, et~al.
\newblock An artificial intelligence framework for automatic segmentation and
  volumetry of vestibular schwannomas from contrast-enhanced t1-weighted and
  high-resolution t2-weighted mri.
\newblock {\em Journal of neurosurgery}, 134(1):171--179, 2019.

\bibitem{simpson2019large}
A.~L. Simpson, M.~Antonelli, S.~Bakas, M.~Bilello, K.~Farahani,
  B.~Van~Ginneken, A.~Kopp-Schneider, B.~A. Landman, G.~Litjens, B.~Menze,
  et~al.
\newblock A large annotated medical image dataset for the development and
  evaluation of segmentation algorithms.
\newblock {\em arXiv preprint arXiv:1902.09063}, 2019.

\bibitem{valindria2018small}
V.~V. Valindria, I.~Lavdas, J.~Cerrolaza, E.~O. Aboagye, A.~G. Rockall,
  D.~Rueckert, and B.~Glocker.
\newblock Small organ segmentation in whole-body mri using a two-stage fcn and
  weighting schemes.
\newblock In {\em International Workshop on Machine Learning in Medical
  Imaging}, pages 346--354. Springer, 2018.

\bibitem{xiang2020learning}
L.~Xiang, G.~Ding, and J.~Han.
\newblock Learning from multiple experts: Self-paced knowledge distillation for
  long-tailed classification.
\newblock In {\em European Conference on Computer Vision}, pages 247--263.
  Springer, 2020.

\bibitem{zela2019understanding}
A.~Zela, T.~Elsken, T.~Saikia, Y.~Marrakchi, T.~Brox, and F.~Hutter.
\newblock Understanding and robustifying differentiable architecture search.
\newblock {\em arXiv preprint arXiv:1909.09656}, 2019.

\bibitem{zhang2021understanding}
C.~Zhang, S.~Bengio, M.~Hardt, B.~Recht, and O.~Vinyals.
\newblock Understanding deep learning (still) requires rethinking
  generalization.
\newblock {\em Communications of the ACM}, 64(3):107--115, 2021.

\bibitem{zhang2020exploring}
D.~Zhang, G.~Huang, Q.~Zhang, J.~Han, J.~Han, Y.~Wang, and Y.~Yu.
\newblock Exploring task structure for brain tumor segmentation from
  multi-modality mr images.
\newblock {\em IEEE Transactions on Image Processing}, 29:9032--9043, 2020.

\bibitem{zhang2021automatic}
D.~Zhang, J.~Zhang, Q.~Zhang, J.~Han, S.~Zhang, and J.~Han.
\newblock Automatic pancreas segmentation based on lightweight dcnn modules and
  spatial prior propagation.
\newblock {\em Pattern Recognition}, 114:107762, 2021.

\bibitem{zhou2019prior}
Y.~Zhou, Z.~Li, S.~Bai, C.~Wang, X.~Chen, M.~Han, E.~Fishman, and A.~L. Yuille.
\newblock Prior-aware neural network for partially-supervised multi-organ
  segmentation.
\newblock In {\em Proceedings of the IEEE/CVF International Conference on
  Computer Vision}, pages 10672--10681, 2019.

\bibitem{zhou2017fixed}
Y.~Zhou, L.~Xie, W.~Shen, Y.~Wang, E.~K. Fishman, and A.~L. Yuille.
\newblock A fixed-point model for pancreas segmentation in abdominal ct scans.
\newblock In {\em International conference on medical image computing and
  computer-assisted intervention}, pages 693--701. Springer, 2017.

\end{thebibliography}


\end{document}